\documentclass[fleqn,10pt]{wlscirep}
\usepackage[utf8]{inputenc}
\usepackage[T1]{fontenc}
\usepackage{multirow}
\usepackage{array}
\usepackage{booktabs}
\usepackage{amsmath, amssymb}
\usepackage[table]{xcolor}
\usepackage{bbm}
\usepackage{xcolor}
\usepackage{algorithm}
\usepackage{algpseudocode}
\usepackage{caption}
\usepackage{subcaption}
\usepackage[percent]{overpic}

\usepackage{float}
\usepackage{graphicx}

\usepackage{newtxtext,newtxmath}  

\usepackage{appendix}
\title{Biosignal Fingerprinting: A Cross-Modal PPG-ECG Foundation Model}

\author[1,2,+]{Zhangdaihong Liu}
\author[3,+]{Chang Liu}
\author[2]{Fenglin Liu}
\author[1,2]{Yixuan Chen}
\author[3,*]{Yang Yang}
\author[1,2]{David~A.~Clifton}
\author[1,*]{Xiao Gu}
\affil[1]{Department of Engineering Science, University of Oxford, Oxford OX3 7DQ, UK}
\affil[2]{Oxford Suzhou Centre for Advanced Research, Suzhou, 215123, China}
\affil[3]{School of Public Health, Shanghai Jiao Tong University, Shanghai, China}
\affil[*]{emma002@sjtu.edu.cn; xiao.gu@eng.ox.ac.uk}

\affil[+]{these authors contributed equally to this work}

\begin{abstract}
Cardiovascular disease remains the leading cause of global mortality, yet scalable cardiac monitoring is hindered by the gap between the diagnostic richness of electrocardiography (ECG) and the ubiquity of wearable photoplethysmography (PPG). Bridging this gap requires representations that are compact, transferable across modalities and devices, and deployable without task-specific retraining. Here we introduce biosignal fingerprints: compact latent representations of cardiovascular state derived from a cross-modal foundation model, the Multi-modal Masked Autoencoder (M2AE), trained on {over 3.4 million} paired ECG and PPG signals. M2AE integrates modality-specific encoders with a shared bottleneck and dual decoders, jointly optimized using reconstruction and cross-modal contrastive objectives, yielding generalizable fingerprints that retain both intra- and inter-modality features. Like a biometric fingerprint, these representations uniquely encode an individual's cardiovascular state in a form that is modality-agnostic, privacy-preserving, and reusable across diverse clinical tasks without exposing raw waveform data or requiring model retraining. Across {7 downstream tasks}, spanning cross-modal reconstruction, cardiovascular disease classification, hypertension detection, mortality prediction, and demographic inference, biosignal fingerprints achieve competitive or superior performance compared to leading domain-specialist foundation models in fully frozen settings, including an {AUROC of 0.974} for five-class CVD classification and {0.877} for hypertension detection, with a maximum improvement of {27.7\%} in AUROC across 5 classification tasks compared to the baseline models. Critically, strong performance is maintained even when only a single modality is available, enabling deployment in the resource-constrained, single-sensor environments typical of real-world wearable monitoring, with direct implications for continuous cardiovascular monitoring across both clinical and consumer health settings.

\end{abstract}
\begin{document}

\flushbottom
\maketitle
%
%
\thispagestyle{empty}

\section*{Introduction}
Wearable health technologies are rapidly transforming the landscape of biomedical monitoring and digital health. From consumer-grade fitness trackers to medical devices, wearable biosensors have become integral to capturing multi-modal physiological data in real time, enabling longitudinal assessment across the multistage trajectory of disease. This expands from asymptomatic risk stratification and early detection to dynamic monitoring of disease progression, treatment response, and long-term outcomes in non-clinical environments. Despite this proliferation of data, a fundamental bottleneck remains: there is no general-purpose representation of physiological state that is compact enough for on-device deployment, transferable across modalities and tasks, and robust to the missing or heterogeneous data that characterizes real-world wearable settings. To address this, we propose a multi-modal biosignal foundation model that distills paired ECG and PPG signals into biosignal fingerprints — compact, modality-agnostic latent representations of cardiovascular state that can be stored, transferred, and reused across tasks without retraining..

Electrocardiography (ECG) and photoplethysmography (PPG) are the two most widely used modalities for cardiac monitoring in both clinical and consumer settings. ECG, particularly the 12-lead configuration, offers high temporal and spatial resolution for capturing cardiac electrical activity and remains indispensable for diagnosing myocardial infarction, arrhythmias, and structural abnormalities \cite{clifford2017af, charlton2022wearable,gu2025label}, with recent deep learning advances extending its utility to single-lead wearable configurations \cite{ribeiro2020automatic}. However, ECG acquisition typically requires periodic clinical visits or active patient engagement, limiting its suitability for long-term, unobtrusive monitoring \cite{neri2023electrocardiogram}.
PPG, by contrast, is a low-cost optical technique that infers blood volume changes in peripheral vasculature and has been seamlessly integrated into smartwatches and rings \cite{kim2023photoplethysmography, loh2022application, pereira2020photoplethysmography}, making it ideal for continuous, passive monitoring at scale. Nevertheless, PPG measures downstream cardiovascular signals and lacks the sensitivity to directly detect complex cardiac events \cite{pereira2020photoplethysmography}, while remaining susceptible to motion artifacts and ambient light interference. 

Given these complementary characteristics, combining ECG and PPG holds great promise, leveraging both the specificity of ECG and the ubiquity of PPG. {Prior studies have shown that ECG and PPG share cardiovascular information supporting applications from cardiac abnormality detection \cite{liu2023intelligent} to vital sign estimation \cite{gu2025sensing}. Yet in free-living wearable settings, simultaneous acquisition of both modalities is often impractical, since most devices capture only one signal type during deployment. This makes it important to learn representations that can exploit the shared cardiovascular structure between ECG and PPG during pretraining while remaining useful when only a single modality is available at inference.} 

Foundation models provide a promising way towards such direction. The success of foundation models in natural language processing \cite{brown2020language} and computer vision \cite{he2022masked} has demonstrated that large-scale pre-training on diverse datasets can yield powerful, general-purpose representations \cite{bommasani2021foundation}. These models have already shown promise in healthcare applications involving language and image modalities \cite{mathew2024foundation, kim2024health, vaid2023foundational}. Recent work has extended these paradigms to physiological time series \cite{chen2024eegformer, ding2024siamquality, abbaspourazad2024large}, such as ECG and other biosignals \cite{luo2025foundationmodelmultivariatewearable, ding2024siamquality, gu2025sensing, gu2025survey}. Notably, Gu et al. \cite{gu2025sensing} proposed a multimodal cardiac foundation model trained on heterogeneous ECG, PPG, and textual data to support transfer across scenarios, devices, and input configurations. This line of work highlights the promise of large-scale pretraining for cardiac sensing. Our study addresses a complementary problem: learning compact shared representations from paired ECG and PPG that explicitly support cross-modal alignment, reconstruction, and robust use, especially when only one modality is available at inference.

\begin{figure}[ht]
\centering
\includegraphics[width=0.9\linewidth]{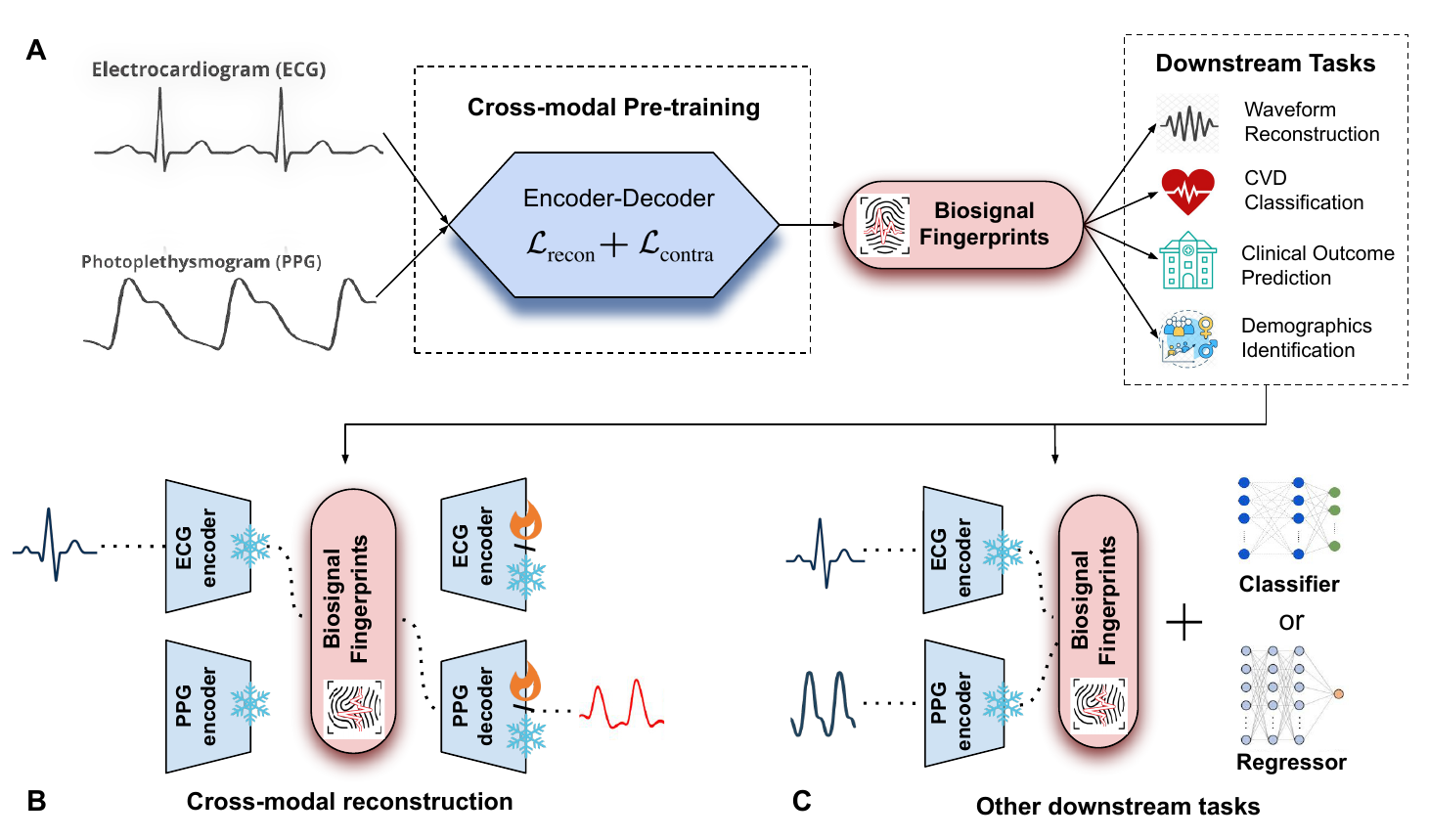}
\caption{\textbf{Analysis pipeline overview.} \textbf{A}: M2AE cross-modal pre-training pipeline. Specifically, the sources of input are paired single-lead (lead II) ECG and PPG waveform segments acquired simultaneously, fed jointly into an encoder-decoder architecture optimized end-to-end using reconstruction and cross-modal contrastive objectives. This training yields biosignal fingerprints at the bottleneck, which serve as compact, modality-agnostic representations of cardiovascular state and support a wide range of downstream applications. \textbf{B}: Illustration of how biosignal fingerprints are utilized in cross-modal reconstruction pipeline under two settings. In the fully frozen setting, both the encoders and decoders are kept frozen; in the decoder fine-tuning setting, the encoders remain frozen to fix fingerprint generation while only the corresponding decoders are made trainable to reconstruct the target modality waveform. \textbf{C} Illustration of the pipeline for other downstream tasks, including CVD classification, clinical outcome prediction, and demographics identification, where biosignal fingerprints extracted from frozen encoders are passed to a lightweight classifier or regressor. Specific model and training details are explained in the Methods Section. An optional single-modal warm-start was evaluated as an ablation and is reported in the Appendix \ref{app.warmstart}.}
\label{fig:pipline}
\end{figure}

To address these challenges, we present the first cross-modal foundation model trained on millions of ECG and PPG signals, and introduce biosignal fingerprints as a new paradigm for physiological representation learning. Our framework is based on a masked autoencoder design for time series \cite{yang2023biot}, building on prior work demonstrating the effectiveness of contrastive learning for cardiac signal representation \cite{kiyasseh2021clocs, geng2022multimodal}. The resulting Multi-modal Masked Autoencoder (M2AE) (details explained in Fig.~\ref{fig:pipline} A and the Methods Section) consists of dual modality-specific encoders and decoders fused at a central bottleneck, jointly optimized using reconstruction and cross-modal contrastive objectives. Biosignal fingerprints are extracted from this bottleneck and capture core information shared by ECG and PPG, retaining both intra- and inter-modality features in a compact, transferable form.

Crucially, these fingerprints are designed to be general-purpose: once computed, they can be applied to a wide range of downstream tasks, including cross-modal reconstruction, CVD classification, mortality and hypertension prediction, and demographic inference, without task-specific fine-tuning of the foundation model, and without requiring both modalities to be present at inference (Fig.~\ref{fig:pipline} Panels B and C). By abstracting high-dimensional, heterogeneous waveforms into stable, reusable biosignal fingerprints, we establish a new paradigm for wearable biosignal analysis — one that accommodates missing modalities, enhances cross-device interoperability, maximizes the utility of limited sensing environments, and enables privacy-preserving, on-device analytics at scale.

\section*{Results}\label{sec:results}

To evaluate the effectiveness of our pre-trained foundation models, particularly the biosignal fingerprints derived from the cross-modal foundation model, we designed a comprehensive suite of downstream tasks. These tasks include time series reconstruction, regression, binary classification, and multi-class classification. An overview of the task types and corresponding datasets is provided in Table \ref{tab:task_overview}. Specifically, we included a waveform-based cross-modal reconstruction task to demonstrate the foundation model’s ability to learn generalizable representations across modalities and modality-agnostic imputation. To assess modality-specific tasks, we conducted several cardiovascular disease (CVD) categorizations under conditions of missing modality input as well as multi-modal input, evaluating the model’s intrinsic diagnostic capabilities. We further implemented two general clinical outcome-related tasks, mortality prediction and hypertension identification, both as binary classification tasks. Finally, age regression and gender classification were included as auxiliary tasks to probe the extent to which the learned fingerprints encode demographic and physiological attributes.

To rigorously assess the representational quality of the biosignal fingerprints, all downstream tasks were conducted using a linear probing approach \cite{chen2020big}, in which the encoders and biosignal fingerprints were kept frozen. For classification and regression tasks, the biosignal fingerprints were used as inputs to an XGBoost classifier or regressor (see Fig.~\ref{fig:pipline}C). 

\begin{table}[ht]
\centering
\caption{\textbf{Overview of tasks and their corresponding datasets.}  For binary classification tasks, the Label Ratio column indicates the proportion of positive samples: male for gender prediction, and deceased and hypertensive for mortality prediction and hypertension identification, respectively. For the 3-class task, the Label Ratio is Normal : Atrial Fibrillation : Premature Atrial/Ventricular Contractions.}
\begin{tabular}{llllcc}
\toprule
\textbf{Task Name} & \textbf{Task Type} & \textbf{Dataset Name} & \textbf{Modality} &\textbf{Dataset Size} & \textbf{Label Ratio} \\
\midrule
Intra- and Cross-modal Reconstruction & Reconstruction & PulseDB &ECG+PPG & 430k & - \\
\hline
\multirow{2}{*}{CVD Classification}                    
& 5-class  & CODE15 &ECG & 26.8k & balanced \\
& 3-class  & SIMBAND  &PPG & 876& 15:4:3\\
\hline
Mortality Prediction                  & Binary         & CODE15 &ECG & 16.9k & 49.28\% \\
Hypertension Detection           & Binary         & MIMIC-III &ECG+PPG & 51.4k & 51.79\% \\
Age Identification                    & Regression                   & MIMIC-III &ECG+PPG & 49.4k & -  \\
Gender Identification    & Binary         & MIMIC-III &ECG+PPG & 51.4k & 53.39\% \\
\bottomrule
\end{tabular}
\label{tab:task_overview}
\end{table}

We conducted method comparisons against several state-of-the-art foundation models, which have published codes and models, including ECG-FM \cite{mckeen2024ecgfmopenelectrocardiogramfoundation}, NormWear \cite{luo2025foundationmodelmultivariatewearable}, PaPaGei \cite{pillai2025papageiopenfoundationmodels}, Chronos \cite{ansari2024chronos}, and MOMENT \cite{goswami2024moment}, and SimCLR \cite{chen2020simclr} where applicable based on the data modalities of the task. For a fair comparison, all baseline encoders were used in a frozen state, and their extracted representations were employed as input features to downstream classifiers or regressors. The comparison evaluation framework is illustrated in Fig.~\ref{fig:pipline}C. Furthermore, we conducted a qualitative assessment, applying UMAP \cite{mcinnes2018umap} to visualize the biosignal fingerprints as well as the embeddings extracted from the baseline models for different classification tasks \cite{becht2019dimensionality}. 

\begin{table}[]
\centering
\caption{\textbf{Training dataset overview.} The last three columns indicate whether each dataset is used in (i) the single-modal warm-start ablation (ECG-only / PPG-only; Appendix \ref{app.warmstart}) and (ii) the main cross-modal foundation model pre-training (paired ECG+PPG). The bottom row shows the total size of the training data for each phase. The unit \textit{m} denotes millions, and \textit{k} denotes thousands. For each training phase, 10\% of the data was reserved as a test set for downstream evaluation, while another 10\% was used as a validation set for hyperparameter tuning.}\label{tab:data}
\begin{tabular}{lccccc}
\toprule
\textbf{Dataset}   & \textbf{Size} & \textbf{Sampling Frequency} & \textbf{ECG-only (ablation)} & \textbf{PPG-only (ablation)} & \textbf{Cross-modality} \\ \midrule
CODE-100\% & 1.56m                & 400Hz         &\checkmark      & &\\
PTB-XL & 21.4k &500Hz    &\checkmark  & &\\
MIMIC-III WDB & 1.95m                & 125Hz       &\checkmark    &\checkmark    &\checkmark  \\ 
VitalDB  &1.15m &500Hz  &\checkmark &\checkmark     &\checkmark  \\ 
\hline
\hline
\textbf{Total size} & & & \textbf{4.26m} &\textbf{3.41m} &\textbf{3.41m}\\\bottomrule
\end{tabular}
\end{table}

To develop robust cross-modal foundation models and high-quality biosignal fingerprints, we directly pretrained M2AE on paired ECG and PPG segments (10 seconds each). Specifically, we used 3.41 million paired segments sourced from MIMIC-III \cite{johnson2016mimiciii} and VitalDB \cite{wang2023pulsedb}, which provide simultaneous ECG and PPG recordings and are therefore suitable for cross-modal representation learning.

The characteristics for each dataset used are summarized in Table \ref{tab:data}. ECG-only datasets such as CODE and PTB-XL \cite{wagner2020ptbxl} are not suitable for cross-modal pre-training and were therefore used only for downstream evaluations that require ECG inputs.

\begin{figure}[h!]
\centering
\includegraphics[width=0.33\linewidth]{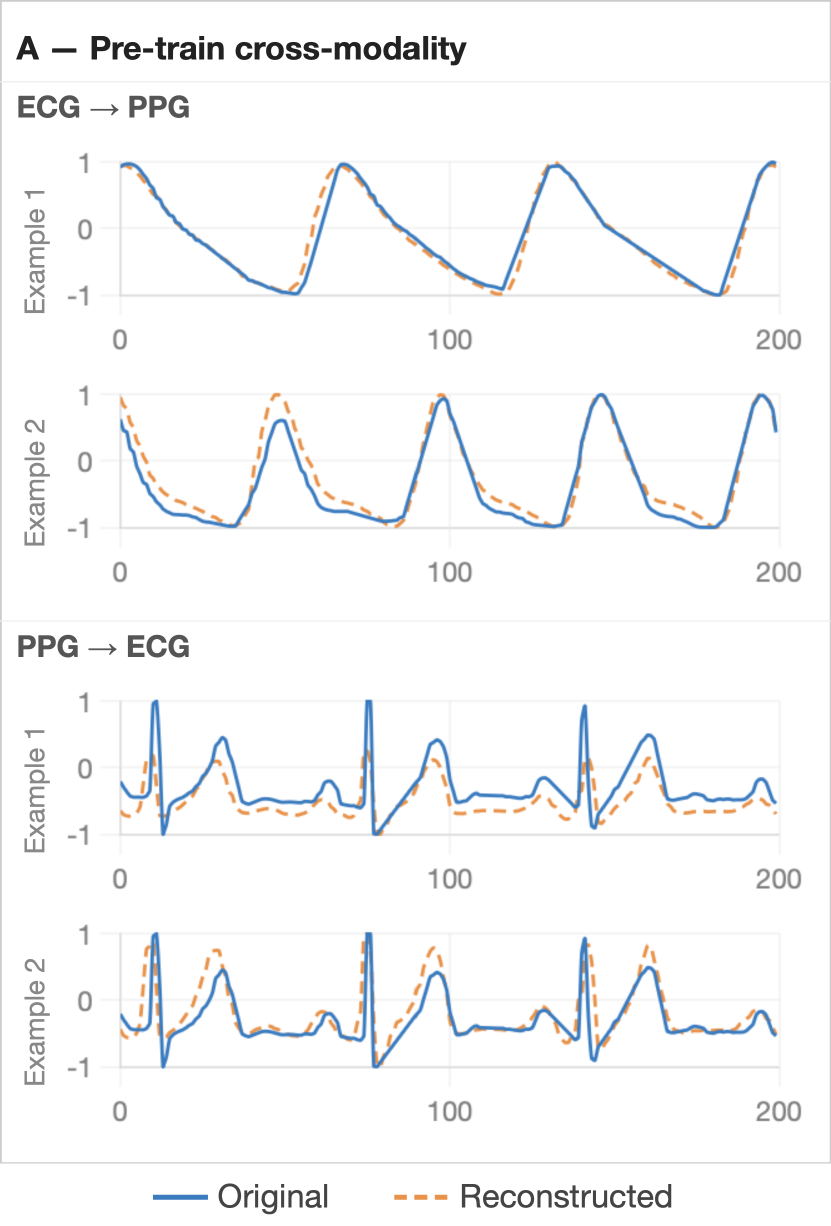}
\includegraphics[width=0.33\linewidth]{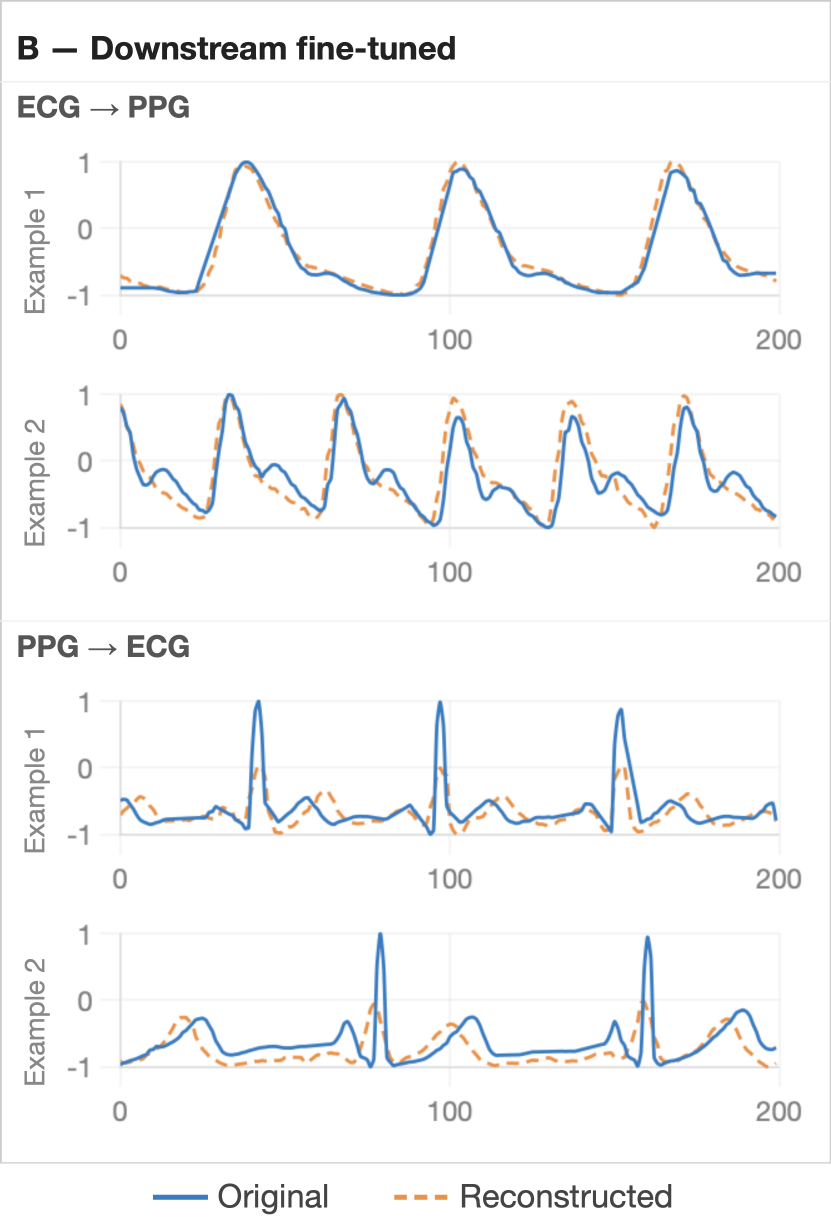}
\includegraphics[width=0.33\linewidth]{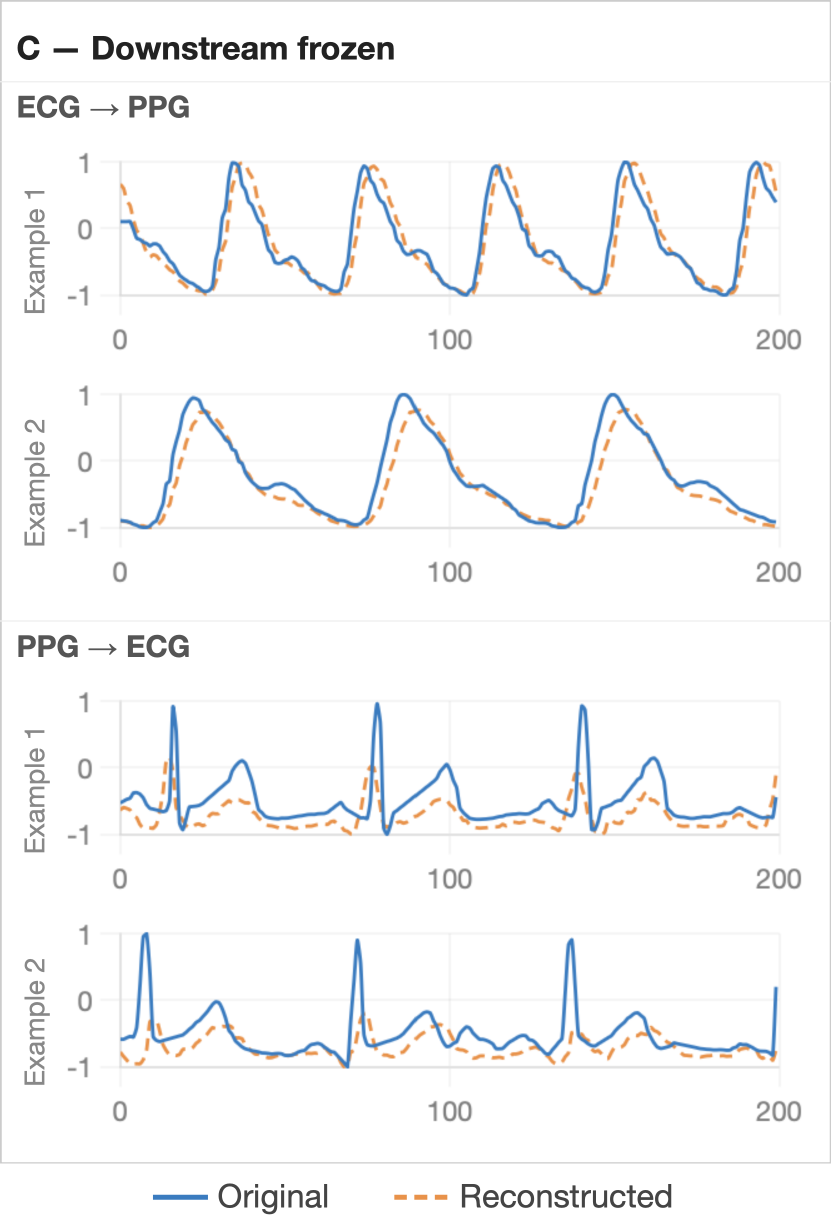}
\caption{\textbf{Visualization of the original and reconstructed biosignals}. \textbf{A:} pre-training with cross-modal reconstruction objectives, \textbf{B:} downstream fine-tuning, and \textbf{C:} downstream inference with frozen encoder and decoder. Each panel shows two representative examples for ECG-to-PPG reconstruction (top rows) and PPG-to-ECG reconstruction (bottom rows). Blue lines denote the original signal; orange dashed lines denote the model reconstruction.} 
\label{fig:singleFM}
\end{figure}

We optimized the cross-modal foundation model end-to-end using reconstruction and cross-modal contrastive objectives. Qualitative reconstruction examples during cross-modal pre-training are shown in Fig.~\ref{fig:singleFM}A.

\subsection*{Biosignal Fingerprints Demonstrate Robust Cross-Modal Reconstruction Capabilities}

The cross-modal foundation model was trained directly from scratch on paired ECG and PPG segments, with modality-specific encoders producing latent embeddings that are fused into a unified set of biosignal fingerprints (details can be found in the Methods Section). These shared fingerprints are then passed through their respective decoders to independently reconstruct the original ECG and PPG waveforms. 

In addition to the reconstruction loss, a cross-modal contrastive loss was introduced to harmonize the updates of both encoders and decoders, thereby facilitating the learning of cross-modality representations. The natural positive pairs for contrastive learning consisted of ECG and PPG segments collected simultaneously from the same individual. To prevent the model from relying solely on temporal proximity, the definition of positive pairs was extended to include: (i) segments from the same modality and individual but collected at different times, and (ii) segments from the same individual across different modalities and acquisition times. The total loss was the weighted sum of the reconstruction and contrastive loss. The details of data augmentation are illustrated in Appendix \ref{app:algo}, and the details of cross-modal contrastive learning are explained in the Methods Section.

The biosignal fingerprints demonstrated a strong capacity to reconstruct the original waveform across modalities during training (Fig.~\ref{fig:singleFM}A). To evaluate their performance in downstream cross-modal reconstruction, we employed two settings (Fig.~\ref{fig:pipline} B and C): (i) decoder fine-tuning, where the encoders were frozen to fix fingerprint generation while only the decoders remained trainable; and (ii) fully frozen inference, where the entire cross-modal network was kept frozen, and the pre-trained biosignal fingerprints were directly used for waveform reconstruction. 
As expected, decoder fine-tuning led to reduced reconstruction error, achieving MAEs of $0.411$ (ECG-to-PPG) and $0.566$ (PPG-to-ECG). Importantly, even under the fully frozen setting, the biosignal fingerprints achieved competitive MAEs of $0.417$ (ECG-to-PPG) and $0.577$ (PPG-to-ECG), suggesting that the learned representations preserve sufficient morphological detail for waveform recovery without any additional training. This is clinically significant: in wearable settings where ECG acquisition is intermittent or impractical, the ability to reconstruct ECG waveforms from PPG alone using frozen fingerprints, could enable passive, continuous estimation of cardiac electrical activity from optical sensors, supporting arrhythmia screening and cardiac monitoring in scenarios where electrode-based recording is unavailable.
Qualitative visualizations of reconstructed signals under both settings are provided in Fig.~\ref{fig:singleFM}B and Fig.~\ref{fig:singleFM}C, respectively. Notably, all evaluations were performed on independent test sets to ensure unbiased and robust performance assessment.
The loss levels for the cross-modal foundation model during the pre-training are reported in Appendix Table \ref{apptab:hypetune}.



\subsection*{Biosignal Fingerprints Achieve Strong Performance in CVD-Related Tasks}
The biosignal fingerprints demonstrate robust performance in modality-specific tasks, even when only one modality is available. We adopted two publicly available datasets with CVD annotations: CODE15 \cite{ribeiro2021code15} for 5-class classification using ECG and SIMBAND for 3-class classification using PPG. Dataset details are listed in Table \ref{tab:task_overview}. The CVD labels in CODE15 include ``1dAVb'' (1st degree AV block), ``SB'' (sinus bradycardia), ``AF'' (atrial fibrillation), ``ST'' (sinus tachycardia), and normal signals; the SIMBAND labels are ``NSR'' (normal sinus rhythm), ``AF'' (atrial fibrillation), and ``PAC/PVC'' (premature atrial/ventricular contractions).

To assess the capability of our biosignal fingerprints under realistic single-modality deployment conditions, we conducted experiments using fingerprints generated from ECG only and PPG only. For comparison, we evaluated our model against uni-modal foundation models ECG-FM and PaPaGei for the ECG and PPG only settings, respectively, and multi-modal foundation models NormWear, Chronos and MOMENT across both settings. We report the assessment metrics, including the Area Under the Precision-Recall Curve (AUPRC) and Area Under the Receiver Operating Characteristic Curve (AUROC), in Fig.~\ref{fig:cvd_results}. The full numerical results are shown in Appendix \ref{app.3}.

\begin{figure}[h!]
\centering

\begin{subfigure}[t]{0.49\linewidth}
    \captionsetup{position=top}
    \caption{}
    \includegraphics[width=\linewidth]{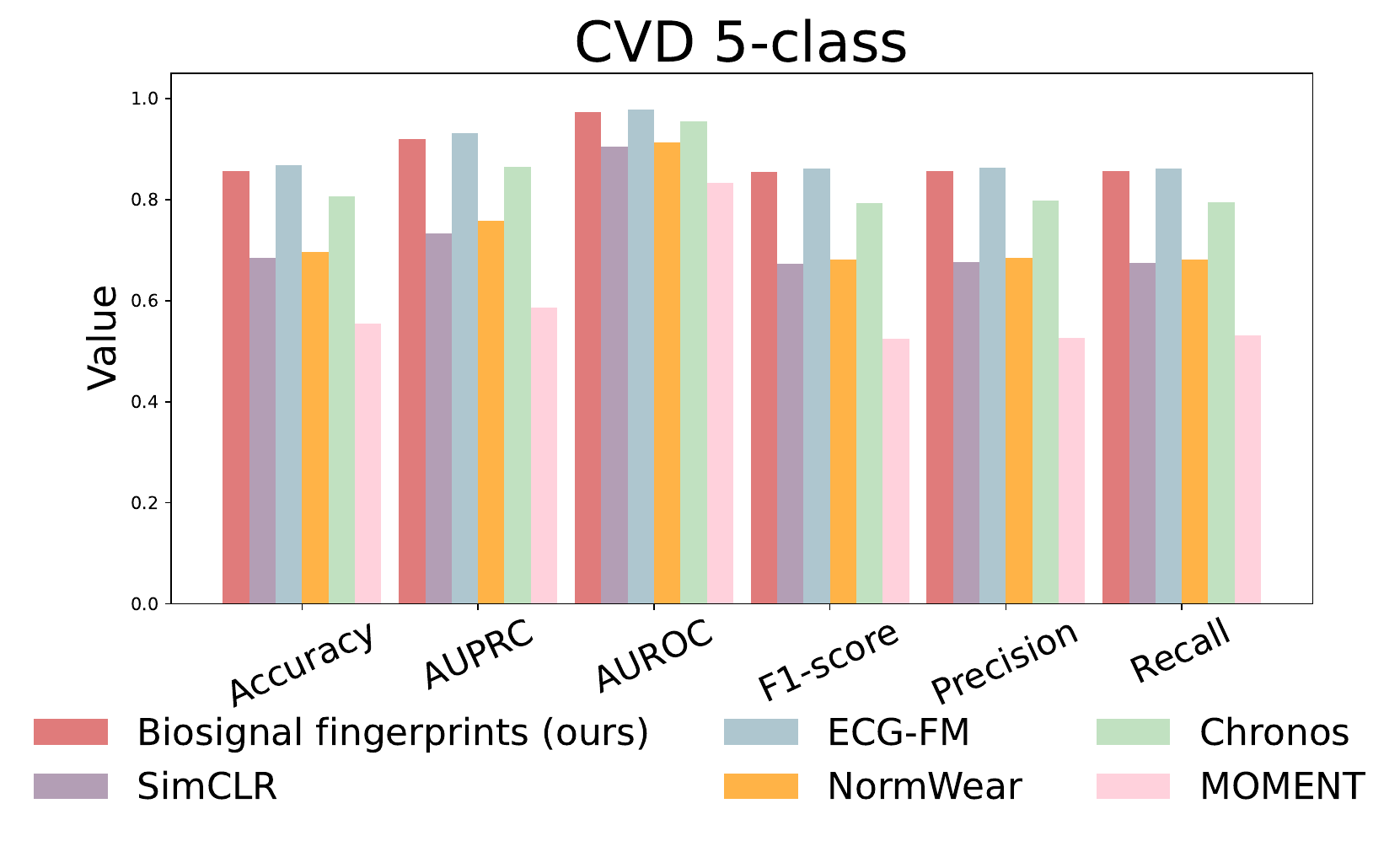}
\end{subfigure}%
\begin{subfigure}[t]{0.49\linewidth}
    \captionsetup{position=top}
    \caption{}
    \includegraphics[width=\linewidth]{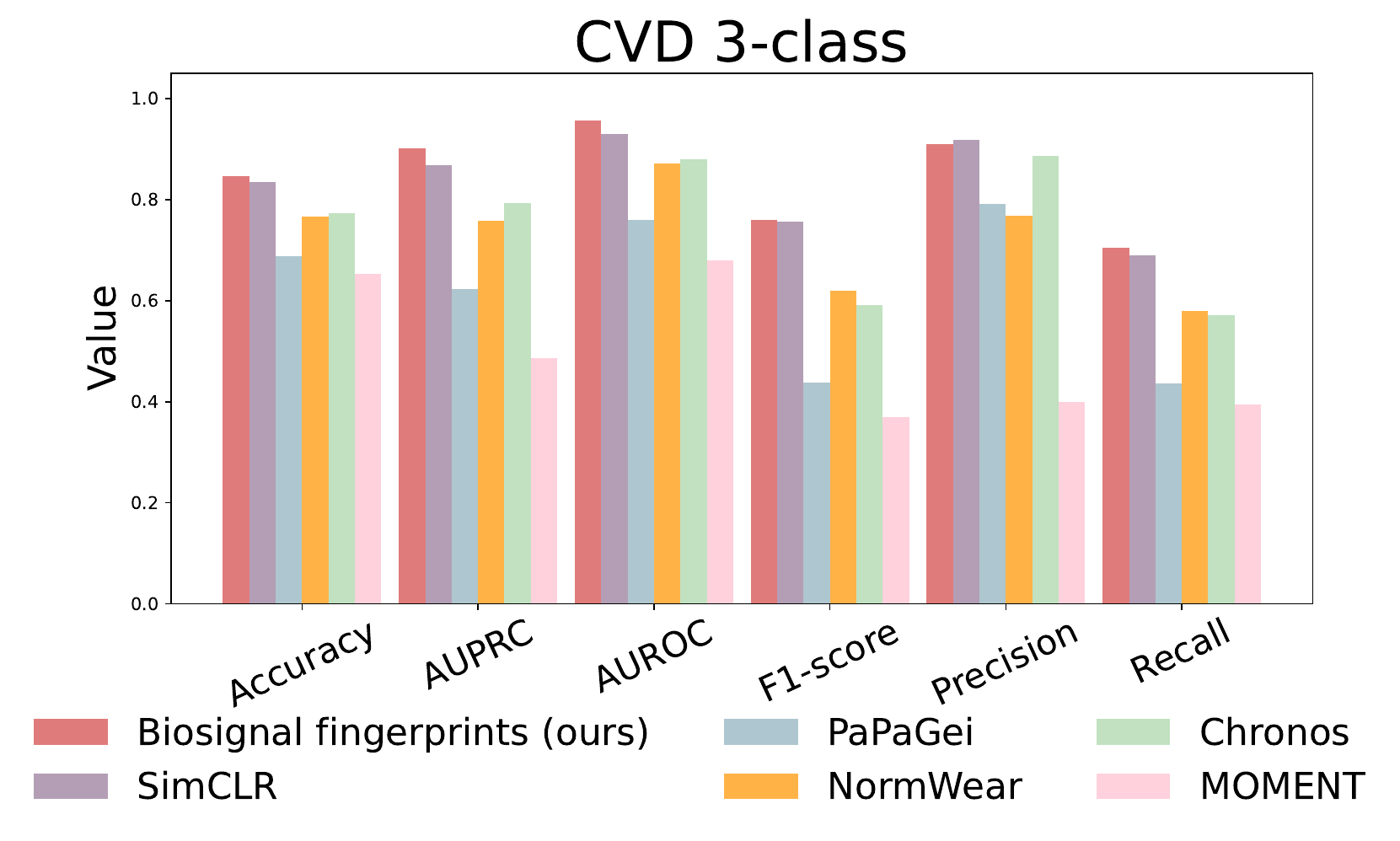}
\end{subfigure}

\vspace{0.5cm}

\begin{subfigure}[t]{\linewidth}
    \captionsetup{position=top}
    \caption{}
    \includegraphics[width=\linewidth]{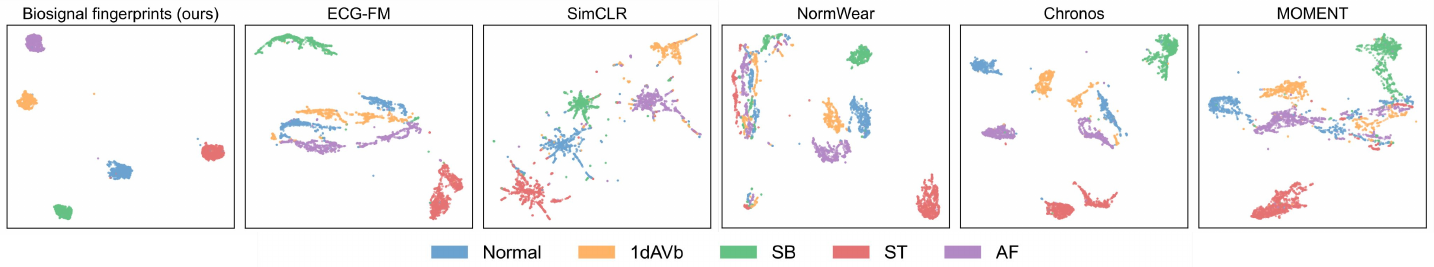}
\end{subfigure}

\caption{\textbf{Results for the CVD-related tasks.} \textbf{A:} Bar plot comparing model performance across six evaluation metrics, including Area Under the Precision-Recall Curve (AUPRC), Area Under the Receiver Operating Characteristic Curve (AUROC), Accuracy, F1-score, Precision, and Recall. The proposed biosignal fingerprints achieve performance comparable to ECG-FM, a foundation model trained exclusively on ECG data, and significantly outperform NormWear, a multi-modal foundation model. \textbf{B:} UMAP projection of embeddings from three models, colored by CVD labels. NormWear exhibits the weakest separation between CVD classes. While ECG-FM embeddings show label-related structure, they produce more fragmented and dispersed clusters. In contrast, the biosignal fingerprints yield more cohesive and clearly delineated clusters. \textbf{C:} UMAP visualization of the latent representation extracted from the 6 foundation models considered. The clusters are colored by the five CVD labels.}
\label{fig:cvd_results}
\end{figure}



\textbf{ECG-only classification.} Our biosignal fingerprints achieved comparable performance with the ECG-specialized foundation model, ECG-FM (Fig.~\ref{fig:cvd_results}A, attaining an AUPRC of 0.920 and AUROC of 0.974 versus 0.931 and 0.978 for ECG-FM respectively, with the largest absolute difference in Accuracy (0.856 vs. 0.869). While this gap is statistically small, it is clinically meaningful to note that our model achieves this using only the ECG encoder, effectively half the M2AE architecture, and without domain-exclusive ECG pretraining. Crucially, our fingerprints significantly outperformed all non-ECG-specialized models, with AUROC improvements of 6.1\% over NormWear and 14.0\% over MOMENT. For a five-class arrhythmia classification task, an AUROC of 0.974 approaches the performance threshold typically associated with clinically deployable automated diagnostic tools, where AUROC values above 0.95 are generally considered to reflect strong discriminative capability.

\textbf{PPG-only classification.} Using PPG as the sole input modality, we performed a multi-class task to identify AF and PAC/PVC. The proposed biosignal fingerprints consistently achieve the best overall performance across the six reported metrics (Fig.~\ref{fig:cvd_results}B. In particular, our method attains AUPRC/AUROC of 0.901/0.956, significantly outperforming the second best model (SimCLR) by nearly 4\%. It also improves the threshold-dependent measures simultaneously (Accuracy 0.847, F1-score 0.760, Precision 0.910, Recall 0.705), indicating a favorable precision--recall balance compared with all baseline models. These results suggest that the learned fingerprints transfer effectively to CVD stratification with PPG as input signals only.

Moreover, visualizing the biosignal fingerprints and baseline embeddings (Fig.~\ref{fig:cvd_results}C) reveals that our fingerprints form more cohesive, label-consistent clusters with fewer redundant groupings. By contrast, ECG-FM produces fragmented intra-class structure, while NormWear, Chronos, and MOMENT yield substantially entangled representations in which multiple classes collapse into overlapping or diffuse manifolds. Nevertheless, this tighter cluster geometry does not directly translate to superior downstream classification performance, where biosignal fingerprints trail ECG-FM marginally. This discrepancy likely reflects two factors: first, UMAP is a non-linear projection that prioritizes local structure, and apparent separation in low-dimensional space is not a reliable proxy for discriminability in the full embedding space; second, ECG-FM benefits from domain-exclusive pre-training on ECG data, lending its representations a degree of cardiac-specific granularity that a generalist biosignal model may not fully replicate. That biosignal fingerprints nonetheless approach ECG-FM performance while generalizing across modalities suggests the gap reflects specialization rather than a fundamental representational limitation.

\subsection*{Biosignal Fingerprints Are Strongly Associated with General Clinical Outcomes}

\begin{figure}[h!]
\centering
    \includegraphics[width=1\linewidth]{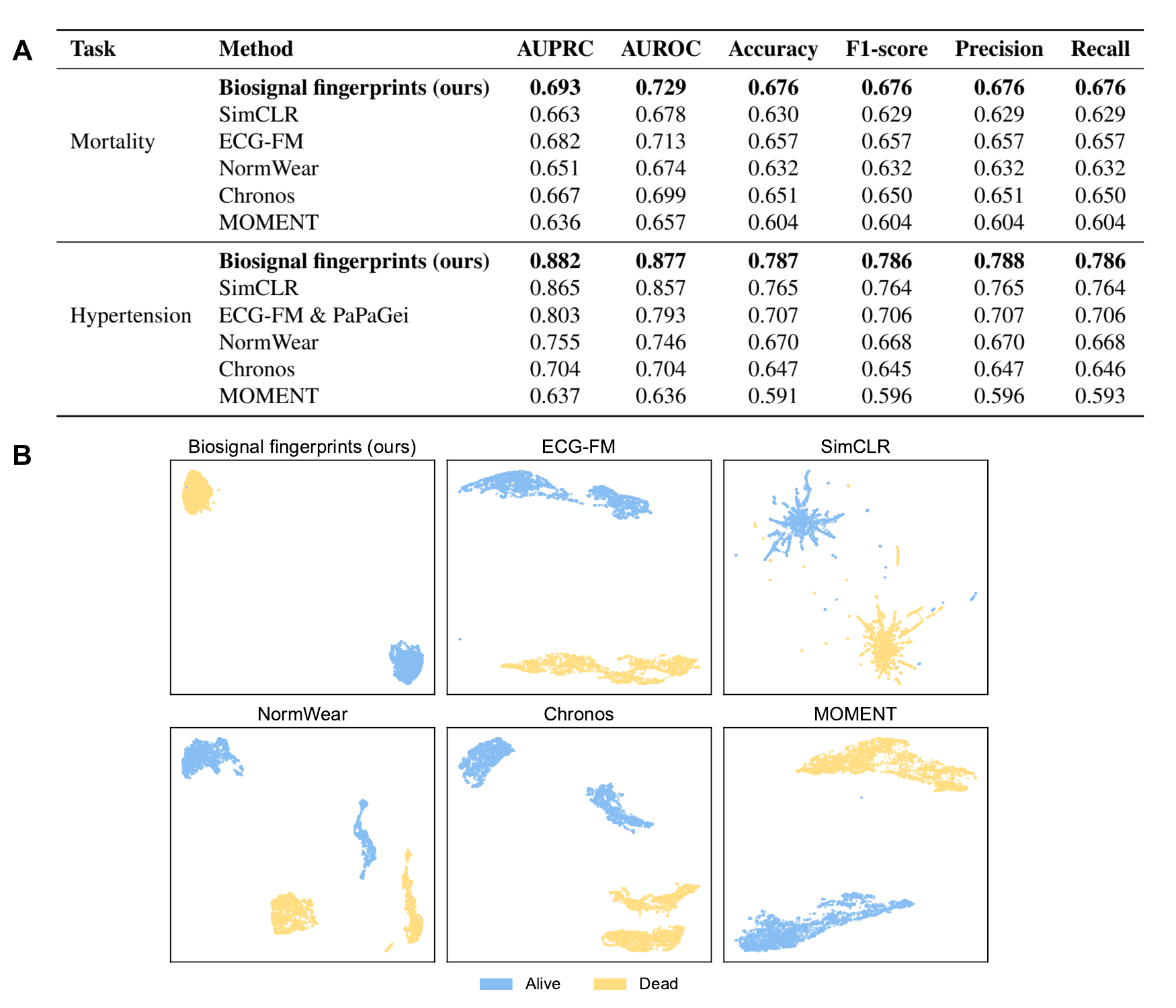}
 \caption{\textbf{Results for general clinical outcomes identification.} \textbf{A:} Quantitative results for mortality and hypertension identification. Both tasks are binary classification tasks. \textbf{B:} UMAP projections of embeddings generated by the six foundation models considered and colored by mortality labels. Our proposed fingerprints demonstrate the clearest separation between alive (blue) and deceased (yellow) individuals and tightest clusters, suggesting a stronger discriminative representation for mortality prediction. In contrast, ECG-FM, MOMENT and SimCLR exhibit more scattered clustering, while NormWear and Chronos display more than two clusters. }
 \label{fig:classresult_umap}
\end{figure}

To evaluate the effectiveness of the proposed biosignal fingerprint in characterizing general clinical outcomes, we leveraged fingerprints generated from the ECG and PPG modalities for two representative tasks: mortality prediction and hypertension detection. Mortality prediction was conducted on CODE15 \cite{ribeiro2021code15} dataset, a large-scale ECG-only dataset with long-term mortality outcomes. Hypertension detection utilized both PPG and ECG modalities from the MIMIC-III Waveform Database Matched Subset and MIMIC-III Clinical Database \cite{johnson2016mimiciii, goldberger2000physionet}. Hypertension was determined based on the International Classification of Diseases (ICD) code “\textit{401 essential (primary) hypertension}” \cite{liu2023patient}. Datasets descriptions are provided in Table \ref{tab:task_overview}. The performance details and comparison against the benchmark models are presented in Fig.~\ref{fig:classresult_umap}A.

Since ECG-FM and PaPaGei are both uni-modal foundation models, we further combined their respective embeddings to form a joint representation for comparison in multi-modal tasks. Our biosignal fingerprints consistently achieved the best performance across all evaluated metrics in both tasks. 
For ECG-only mortality prediction on CODE15, as shown in Fig.~\ref{fig:classresult_umap}A, our biosignal fingerprints achieved the highest discrimination with an AUROC of 0.729 and F1-score of 0.676, improving upon the strongest baseline ECG-FM by $2.2\%$ and $2.9\%$ respectively. While the absolute AUROC of $0.729$ is modest, this reflects the inherent difficulty of long-term mortality prediction from short ECG segments alone, a task where established clinical risk scores such as the Framingham Risk Score and ECG-derived age models typically achieve AUROCs in the 0.70–0.80 range \cite{zacarias2012framingham, li2022comparison, vaid2023foundational}. Our fingerprints therefore operate within a clinically plausible performance envelope for this task, and outperform all competing foundation models including MOMENT by $11.0\%$ in AUROC and $12.0\%$ in F1-score. To qualitatively examine the learned representations, we projected the embeddings of all six models into two dimensions using UMAP \cite{mcinnes2018umap} (Fig.~\ref{fig:classresult_umap}B). Biosignal fingerprints yield the clearest separation between alive and deceased individuals, forming two compact, well-isolated clusters, while all baseline models exhibit considerably more scattered distributions or inter-class overlap. These qualitative results are consistent with the quantitative findings in Fig.~\ref{fig:classresult_umap}A.

On the hypertension identification task, arguably the most clinically impactful result in this work, biosignal fingerprints demonstrated a substantial advantage over all baselines, achieving an AUPRC of 0.882 and AUROC of 0.877. This represents an improvement of 8.5\% in F1-score over the second-best method, ECG-FM \& PaPaGei, and 31.9\% over the weakest baseline, MOMENT. An AUROC of 0.877 for hypertension detection from short wearable biosignal segments is clinically significant: conventional hypertension screening relies on cuff-based blood pressure measurement, which is episodic and subject to white-coat effects, whereas a model operating at this performance level from passively acquired PPG and ECG signals could enable continuous, unobtrusive population-scale screening. The UMAP visualization for this task is appended in Appendix Fig.~\ref{appfig:umap_hyp}.

\begin{figure*}[t!]
\centering
\begin{subfigure}[t]{\linewidth}
    \centering
    \captionsetup{position=top}
    \caption{}
    \includegraphics[width=0.9\linewidth]{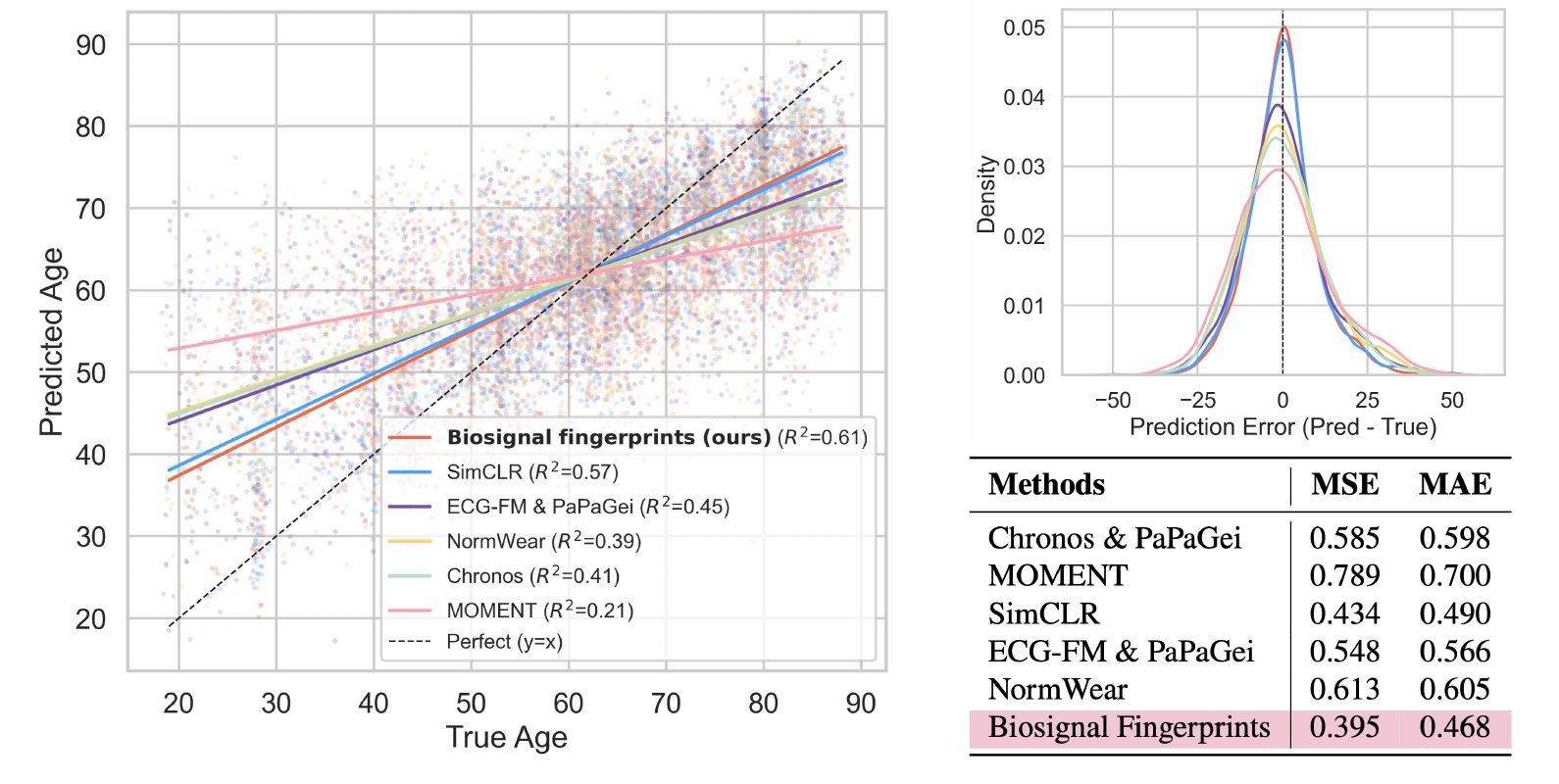}
    \vspace{-10pt}
\label{fig:ageplot}
\end{subfigure}%
\vspace{0.5cm}

     

  \begin{subfigure}[t]{0.4\linewidth}
    \centering
    \captionsetup{position=top}
    \caption{}
    \includegraphics[width=\linewidth]{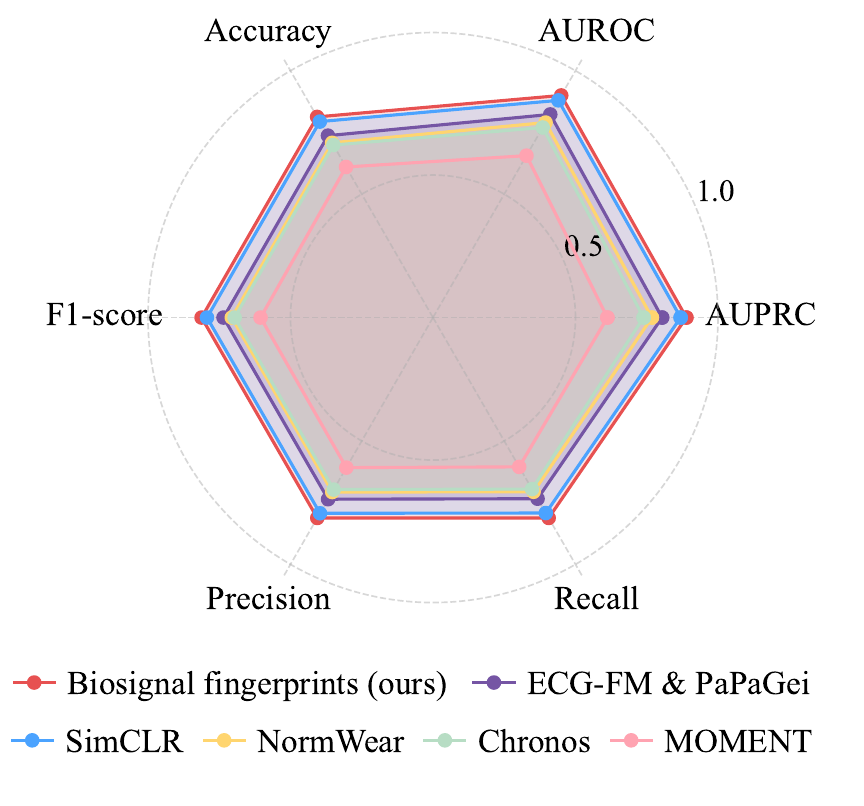}
    \vspace{-10pt}
    \label{fig:radargender}
\end{subfigure}
\hfill
\begin{subfigure}[t]{0.55\linewidth}
    \centering
    \captionsetup{position=top}
    \caption{}
    \includegraphics[width=\linewidth]{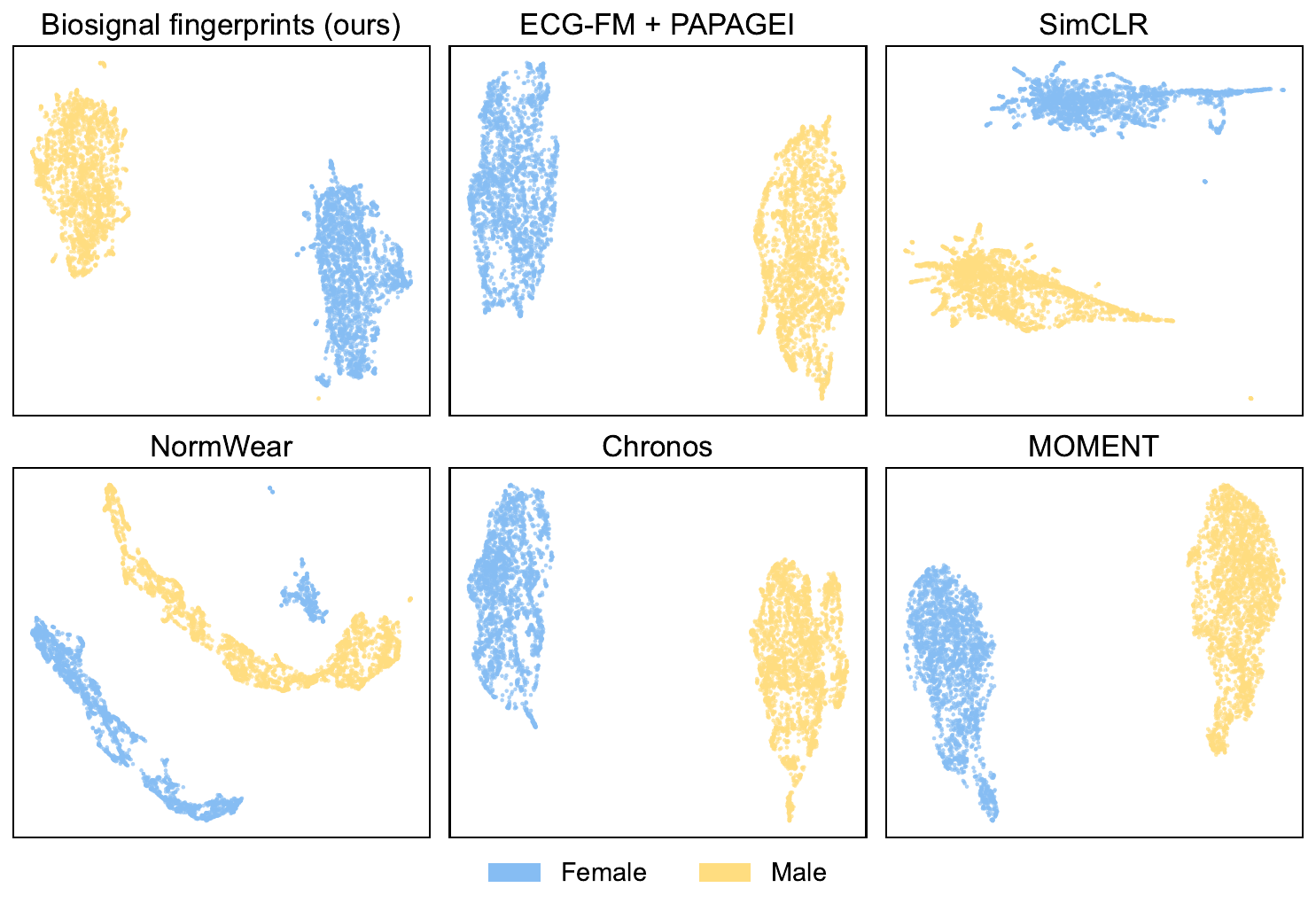}
    \vspace{-10pt}
    \label{fig:umapgender}
\end{subfigure}

 \caption{\textbf{Quantitative and qualitative results for the {Demographics Attribute Inference} tasks.} \textbf{A:} Results for age identification. The left panel shows the scatter plot of predicted ages versus actual chronological ages across all models with their fitted linear model and $R^2$ value. The top right panel shows the predicted error distributions, and bottom right table shows the Mean Absolute Error (MAE) and Mean Squared Error (MSE) for the age prediction task, where lower values indicate better performance. These three illustrations provide a comprehensive view of the superior performance of our proposed biosignal fingerprints against the baseline models. \textbf{B:} Radar chart comparing gender classification performance across six metrics: AUPRC, F1-score, AUROC, Precision, Recall, and Accuracy. The proposed biosignal fingerprints model (red) consistently outperforms all baseline models, with the largest improvement in AURPC being over 45\%, demonstrating superior robustness across all evaluation criteria. \textbf{C:} UMAP projections of biosignal embeddings colored by gender across six models. While all of them exhibit good gender-based separation, our proposed biosignal fingerprints demonstrate tightest clusters (smallest mean within-cluster distance). This suggests that the biosignal fingerprints encode more discriminative and gender-informative representations compared to the baseline models which is in line with the quantitative results.}\label{fig:demo_results}
\end{figure*}

\subsection*{Biosignal Fingerprints Effectively Support Demographic Attribute Inference}

Furthermore, we evaluated the potential of biosignal fingerprints in identifying demographic characteristics, using age and gender, two fundamental demographic indicators, as examples. The MIMIC-III Clinical Database and MIMIC-III Waveform Database Matched Subset \cite{johnson2016mimiciii} were utilized for this task. For the age identification, participants with anonymized ages (over 89 years old) due to the privacy settings of MIMIC were excluded. Dataset details are provided in Table \ref{tab:task_overview}. An XGBoost model was employed for both age and gender prediction. Full implementation details are provided in the Methods Section. For age prediction, as shown in Fig.~\ref{fig:ageplot}, our biosignal fingerprint achieved significantly lower MSE and MAE compared to all baseline models. The predicted ages from our biosignal fingerprints exhibit a markedly stronger correlation with the true ages than those predicted using the baseline embeddings, achieving an $R^2$-value of 0.61. 

Gender classification performance metrics are summarized in Fig.~\ref{fig:radargender} and Appendix \ref{app.3}, where our biosignal fingerprints consistently outperform other state-of-the-art foundation models across all evaluation metrics. The observed performance improvements range from 9.4\% to 15.8\% compared to baseline models. Furthermore, the UMAP visualization in Figure \ref{fig:umapgender} demonstrates that our biosignal fingerprints exhibit superior discriminative capability in separating male and female subjects.

\section*{Discussion}
Our study introduces a novel cross-modal foundation model framework, M2AE, capable of learning robust and generalisable biosignal fingerprints from large-scale paired electrocardiograph (ECG) and photoplethysmograph (PPG) data. These fingerprints demonstrate strong performance across a diverse set of downstream tasks, including waveform reconstruction, cardiovascular disease classification, clinical outcome prediction, and demographic attribute inference. By unifying signal modalities that differ in form, fidelity, and clinical utility, our model sets a new benchmark in the field of physiological representation learning and wearable-based health monitoring.

M2AE builds on a Transformer-based masked autoencoder with modality-specific encoders, a shared bottleneck underpinned by cross-modal contrastive learning, and dual decoders that can reconstruct either or both modalities independently. Unlike prior cross-modal autoencoder work \cite{radhakrishnan2023cross}, the Transformer backbone captures complex temporal dependencies both within and across modalities, while the masking objective promotes compact, corruption-robust representations. Notably, we found that training the cross-modal model from scratch outperforms a warm-start strategy in which single-modal autoencoders are pretrained first and their weights used to initialize the cross-modal model, suggesting that joint cross-modal optimization from the outset is sufficient and preferable, for learning both modality-specific and shared representations simultaneously; full ablation results are reported in Appendix \ref{app.warmstart}. Furthermore, strong downstream performance is maintained even when only one modality is available at inference, underscoring the robustness of the learned fingerprints to the missing-modality scenarios that are common in real-world wearable deployment.

Compared to the latent representations extracted from the leading uni-modal (ECG-FM, PaPaGei) and multi-modal (NormWear) foundation models, our biosignal fingerprints achieve competitive or superior performance across a wide range of downstream tasks and evaluation metrics, especially in more challenging scenarios that are not directly tied to the primary physiological function of the signals, such as hypertension detection and age prediction. This is consistent with recent findings showing that self-supervised pre-training on large-scale ECG datasets can yield representations that generalize effectively across diverse cardiac monitoring tasks, even under limited fine-tuning \cite{lai2023practical}. Notably, these results were achieved in a fully frozen setting, using linear probing without any fine-tuning of the encoders. This demonstrates that our biosignal fingerprints are not only semantically rich and highly transferable, but also eliminate the need for retraining or task-specific tuning of the foundation model. Additionally, the incorporation of contrastive learning across multiple signal modalities and timescales \cite{kiyasseh2021clocs} further enhances the robustness and interpretability of these embeddings, as evidenced by UMAP visualizations and strong generalization to unseen cohorts.

A further consideration concerns the fairness of baseline comparisons. PaPaGei was pre-trained on VitalDB and MIMIC-III Waveform Database, ECG-FM on MIMIC-IV-ECG, and NormWear on several datasets including MIMIC, all of which overlap with the downstream evaluation sets used in this study. Representations from these models may therefore benefit from implicit exposure to the evaluation distribution, potentially inflating their apparent performance. By contrast, our M2AE model enforces subject-level partitioning between pretraining and evaluation splits. The competitive and often superior performance of our biosignal fingerprints under this stricter regime suggests that the reported gains relative to these baselines are likely conservative.

Importantly, the biosignal fingerprints derived from our model serve as a compact, privacy-preserving, and computation-efficient alternative to raw waveform data. By enabling direct application of pre-trained fingerprints in diverse clinical and demographic inference settings, our method reduces both computational overhead and data requirements during implementation, significantly lowering the barrier to applying foundation models in healthcare settings, enabling simpler, faster, and more cost-effective deployment. Moreover, they facilitate deployment in resource-limited settings and support federated or edge-based analytics \cite{xu2021federated}.

Nonetheless, several limitations call for further investigation. First, while our dataset compilation spans millions of signal segments, the cross-modal training remains constrained by the limited availability of high-quality, simultaneously recorded ECG-PPG pairs. Second, while we focused on lead II ECG to reflect typical wearable-device configurations, future extensions, particularly those targeting clinical applications, could benefit from incorporating real-world ECGs and PPGs collected from wearable devices and including additional physiological modalities, such as accelerometry or respiratory signals, to enrich the learned representations and support broader diagnostic utility.

\section*{Conclusions}
In conclusion, our work demonstrates the feasibility and value of cross-modal foundation models for biosignal representation learning. By leveraging self-supervised learning and physiological complementarity, we lay the groundwork for a new class of intelligent, adaptable, and modality-agnostic tools for digital health. As wearable technologies continue to proliferate, the proposed biosignal fingerprinting paradigm holds the promise of enabling earlier detection, broader access, and more personalized insights into cardiovascular and systemic health.

\section*{Methods} \label{sec:method}

\subsection*{Model architecture}
\begin{figure}[h!]
\centering
\includegraphics[width=\linewidth]{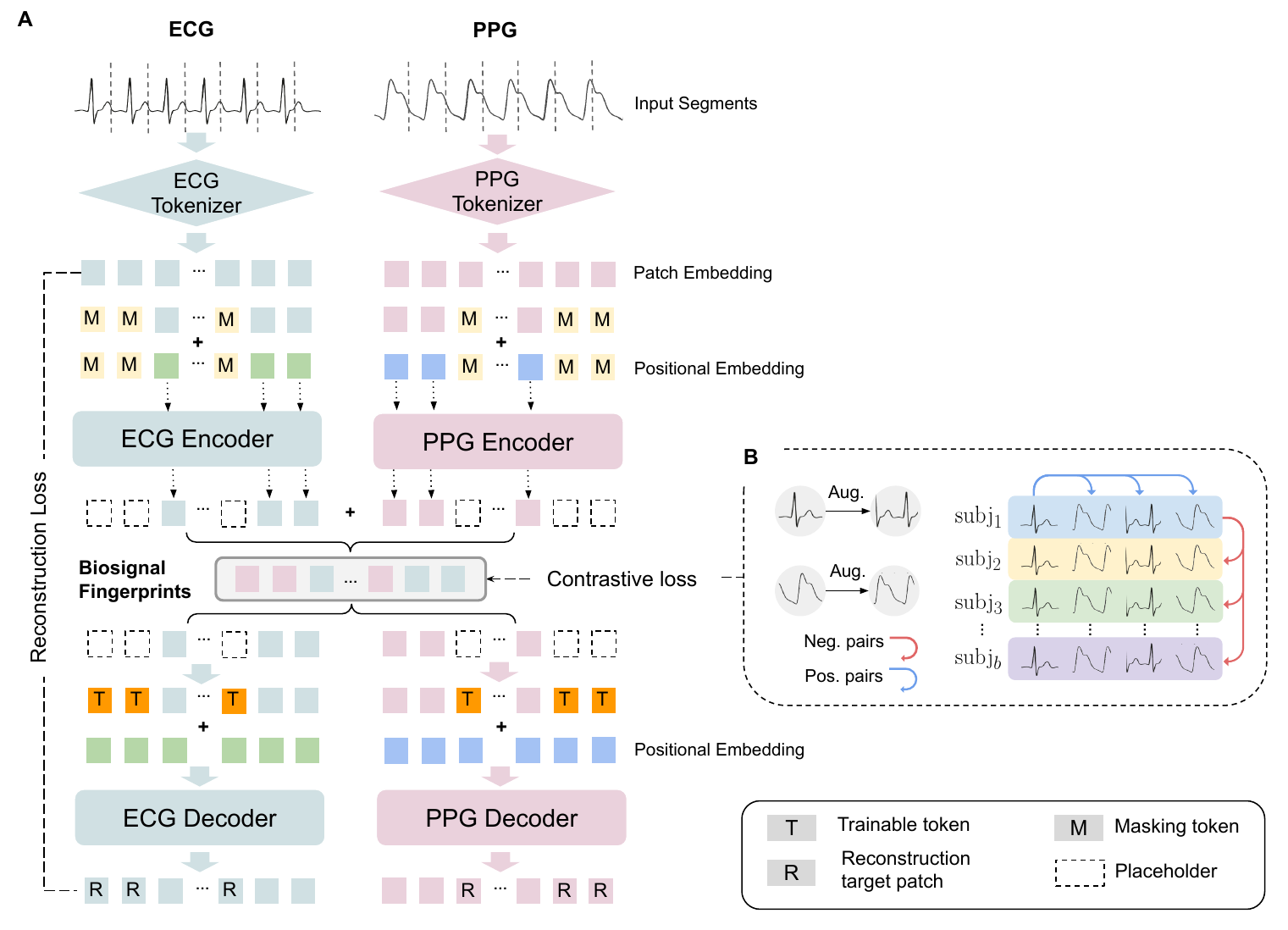}
\caption{\textbf{Method Overview.} Illustration of the cross-modality foundation model. \textbf{A:} This architecture includes dual encoders and decoders for ECG and PPG, which join at the bottleneck through complementary integration to form the biosignal fingerprints. \textbf{B:} Overview of the data augmentation strategy for contrastive learning. The primary positive pair comprises ECG and PPG segments recorded simultaneously from the same individual.}
\label{fig:method}
\end{figure}

The model architecture overview for single-modality and cross-modality networks is illustrated in Fig.~\ref{fig:method}.

Building on recent advances in applying Transformers to biosignals \cite{yang2023biot}, and leveraging the strengths of masked autoencoders for Transformer pre-training, such as the Masked Autoencoder (MAE) for Vision Transformers (ViT) \cite{dosovitskiy2020image}, we propose our cross-modal foundation model architecture: a multi-modal masked autoencoder, referred to as \textbf{M2AE}. This architecture is specifically tailored to time-series data. In essence, M2AE tokenizes the input time series, randomly masks a subset of the resulting patches, and employs a Transformer encoder to learn latent representations that enable reconstruction of the masked patches.

To illustrate this pipeline in detail, let $x_{n,i}^e$ and $x_{n,i}^p$ denote the lead II ECG and PPG signals, respectively, which correspond to the $i$-th sample from the $n$-th subject. Note that a subject may have multiple signal segments. $x_{n,i}^e$ and $x_{n,i}^p$ were modelled identically  by the M2AE framework, therefore, without loss of generality, we represented the signal as $x \in \mathbb{R}^{L}$, with $L=2048$ corresponding to 10-second segment after preprocessing. We first divided $x$ into non-overlapping patches $x_{p} = [p_1, p_2, ..., p_k] \in \mathbb{R}^{k\times s}$, where $k=L//s$ denoting the number of patches and $s=64$ denoting the patch size in this work. Each patch was passed through a \textit{LayerNorm} layer and a \textit{Linear} layer to be represented by a $d$-dimensional embedding, resulting in a tokenized embedding matrix $E_{k\times d}$. For simplicity, we dropped the sample subscript $(n,i)$, and all expressions refer to a single segment from a sample unless otherwise stated. We retained the positional embedding $P^{k\times d}$ from the original ViT to preserve the temporal information. The classification token was omitted, as the pre-training objective was solely reconstruction; thus, only the dense output head was used during training. Notably, the global average embeddings from the were employed as the final fingerprints.

\subsubsection*{Cross-modal training}
The cross-training model, \textbf{M2MAE}, merged the two single-modal models and used the pre-trained weights from these models as initializations. Similar to the single-modal training stage, let $E_{\text{I}}^e$ and $E_{\text{I}}^p$ denote the inputs for ECG and PPG cross-modal encoders, and $r^c_{\text{ecg}}$ and $r^c_{\text{ppg}}$ be the masking ratio for $E_{\text{I}}^e$ and $E_{\text{I}}^p$, respectively. We set $r^c_{\text{ecg}} + r^c_{\text{ppg}} = 1$, allowing them to be merged into a complete set of embeddings at the bottleneck: 
\[
E_{\text{M}}^e[k_i] =
\begin{cases}
    E_{\text{I}}^{e}[k_i], & k_i \in U \quad (\text{unmasked row indices}), \\
    \text{[MASK]}, & k_i \in M \quad (\text{masked row indices}),
\end{cases}
\]

and \[
E_{\text{M}}^p[k_j] =
\begin{cases}
    E_{\text{I}}^{p}[k_j], & k_j \in M \quad (\text{unmasked row indices}), \\
    \text{[MASK]}, & k_j \in U \quad (\text{masked row indices}),\\
\end{cases}
\]
s.t.  \[U, M \in \{1, \dots, k\}, \; U \cup M = \{1, \dots, k\}, \; U \cap M = \varnothing,\]
where $k$ is the total number of patches. Notably, PPG and ECG embeddings were matched at complementary patch indices, allowing them to be merged into a complete set of encoded embeddings at the bottleneck. In fact, to further facilitate the generalizability, for every batch training, we randomly drew  $r^c_{\text{ecg}}\in [0.1,0.9]$, and $r^c_{\text{ppg}} = 1-r^c_{\text{ecg}}$. 

The single merged cross-modal bottleneck embedding is
\begin{equation}
Z^c[k_i] = 
\begin{cases}
f_{\text{enc}}^e(E_{\text{I}}^{e})[k_i], & \text{if } k_i \in U, \\
f_{\text{enc}}^p(E_{\text{I}}^{p})[k_i], & \text{if } k_i \in M,
\end{cases}
\end{equation}
where $f_{\text{enc}}^e$ and $f_{\text{enc}}^p$ are encoders for ECG and PPG, respectively, initialized with the encoders from the single-modal models. The learned $Z^c$, averaged across patches, served as our \textbf{\textit{biosignal fingerprints}} for ECG and PPG after model convergence.

Similar to the single-modal model training, $Z^c$ was then passed through the respective decoders for ECG and PPG, $f^e_{\text{dec}}$ and $f^p_{\text{dec}}$, to obtain the reconstructed patch embeddings $\widetilde{E}^e$ and  $\widetilde{E}^p$. Note that both $f^e_{\text{dec}}$ and $f^p_{\text{dec}}$ were initialized with the decoders from the single-modal training.

The reconstruction loss for the cross-modal training was the sum over both modalities and was calculated only for their respective masked patches:
\begin{equation}\label{eq:recon2}
 \mathcal{L}^c_{\text{recon}} = \frac{1}{|M|} \sum_{k_i \in M} \| E^e[k_i] - \widetilde{E}^e[k_j] \|_2^2   + \frac{1}{|U|} \sum_{k_j \in U} \| E^p[k_j] - \widetilde{E}^p[k_j] \|_2^2.
\end{equation}

\subsubsection*{Cross-modal Contrastive Learning}
The main objective of cross-modal training was to learn a set of latent representations, which we name \textit{biosignal fingerprints}, $Z^c$, that capture common patterns across ECG and PPG while remaining discriminant for different individuals. The learning goals were to draw the fingerprints for the same individual but different biosignal modalities closer together, and to push apart those from different modalities and different individuals. As a self-supervised foundation model framework, we adopted contrastive learning to achieve these goals. 

In this cross-modal setting, we defined \textit{positive pairs} as signals belonging to the same individual and same modality, $(x_{n,i}^e, \; x_{n,j}^e)$ and $(x_{n,i}^p, \; x_{n,j}^p)$, as well as signals belonging to the same individual but different modalities, $(x_{n,i}^e, \; x_{n,i}^p)$. \textit{Negative pairs} were defined as signals belonging to the same modality but different individuals, $(x_{n,i}^e, \; x_{m,j}^e)$ and $(x_{n,i}^p, \; x_{m,j}^p)$, and signals belong to different individual and modalities, $(x_{n,i}^e, \; x_{m,j}^p)$. Notably, for subjects with only one segment of signal, we augmented the signal with a synthesized segment (\textit{e.g.}, adding white noise and time warping to the original segment). During training, all anchor signals were randomly sampled from different subjects to ensure negativity across batch samples. The augmentation was then applied based on the anchor signals. The detailed augmentation strategy is shown in Appendix Algorithm \ref{algo:1}.

To calculate the contrastive loss, let $Z^{\text{ECG}},\; Z^{\text{PPG}},\; Z^{\text{ECG\_aug}},\; Z^\text{{PPG\_aug}} \in \mathbb{R}^{B\times D}$ denote the bottleneck embeddings for the original ECG and PPG and their augmented data, respectively, where $B$ is the batch size and $D$ is dimensionality of the embeddings. 
For an anchor embedding $\mathbf{z}_i^{(v)}$, where the superscript $(v)$ indicates one of the four embedding views (ECG, PPG, ECG\_aug or PPG\_aug), the positives are $\mathbf{z}_i^{(u)}$ for $u\ne v$. Therefore, each anchor embedding has three positive pairs. The negatives are all other samples $\mathbf{z}_j^{(u)}$ with $j\ne i$. 

The InfoNCE loss for an anchor \( \mathbf{z}_i^{(v)} \) is:

\begin{equation}\label{eq:sim}
\mathcal{L}_i^{(v)} =
- \frac{1}{3} \sum_{\substack{u = 1 \\ u \ne v}}^{4}
\log \left(
\frac{
\exp\left( s_{i,i}^{(v,u)} \right)
}{
\sum_{\substack{j=1}}^{N} \sum_{k=1}^{4}
\mathbbm{1}_{(j \ne i \; \text{or} \; k\ne v)} \exp\left( s_{i,j}^{(v,k)} \right)
}
\right),
\end{equation}

where $s_{i,j}^{(v,k)}$ represents the similarity between the embeddings, defined as
\begin{equation}
    s_{i,j}^{(v,u)} = \frac{ \mathbf{z}_i^{(v)} \cdot \mathbf{z}_j^{(u)} }{ \tau }, \quad \text{for} \; i\ne j \; \text{and} \; u\ne v.
\end{equation}
Here, $\tau \in \mathbb{R}_+$ is the temperature parameter. Notably from \ref{eq:sim}, self-similarities are not computed.

The multi-set contrastive loss is:
\begin{equation}\label{eq:contra}
\mathcal{L}_{\text{contrast}} ^c=
\frac{1}{4B} \sum_{i=1}^{B} \sum_{v=1}^{4} \mathcal{L}_i^{(v)}.
\end{equation}

The final loss for the cross-modal training is a weighted sum of the reconstruction loss in \ref{eq:recon2} and the multi-set contrastive loss in \ref{eq:contra}:
\begin{equation}\label{eq:total_loss}
\mathcal{L}^c = \mathcal{L}_{\text{contrast}} ^c + \lambda \mathcal{L}_{\text{recon}} ^c,
\end{equation}
where $\lambda$ is the weighting hyperparameter.

The bottleneck embeddings were extracted after model convergence as the biosignal fingerprint for downstream tasks.

\subsection*{Implementation details}\label{sec:config}
For cross-modal foundation model pre-training, the paired dataset was partitioned into training/validation/test sets in an 80\%:10\%:10\% ratio, respectively. This partitioning was done at the subject level to ensure that no data from the same subject appeared in both training and test sets, thereby preventing data leakage. The training and validation sets were exclusively used for model optimization and hyperparameter tuning. The test set was held out and reserved solely for downstream evaluation tasks and was not accessed during fingerprint generation.

The cross-modal foundation model was trained using PyTorch version 2.4.1, on a computing node equipped with four NVIDIA V100 GPUs, utilizing CUDA version 12.4. Model training was conducted in a distributed manner. PyTorch Lightning version 2.4.0 was employed to facilitate version control and hyperparameter management. For all foundation model and downstream task trainings, we adopted the Adam optimizer to update model parameters. A learning rate scheduler, ReduceLROnPlateau, was employed to reduce the learning rate by a factor of 0.5 if the validation loss failed to improve over two consecutive epochs. To mitigate overfitting, early stopping was implemented based on the validation loss, with patience adjusted according to the specific training objective. Both encoders and decoders were based on Transformer architectures comprising eight attention heads. The encoder has a network depth of six, with an output dimensionality of 768, while the decoder has a network depth of three with an output dimensionality of 256.

For cross-modal training, a batch size of 256 was used, and the learning rate for the optimizer was initialized at $1e-4$. The temperature was set to $0.1$, and the weighting coefficient $\lambda$ in Equation~\ref{eq:total_loss} was empirically fixed at $1$, thereby assigning equal importance to the reconstruction and contrastive loss components. A full list of hyperparameters is provided in Appendix \ref{app.1}. Various hyperparameter configurations were explored, and the corresponding results are presented in the Appendix Table \ref{apptab:hypetune}. 

To robustly evaluate the quality of the learned biosignal fingerprints, we employed a linear probing approach in downstream tasks, wherein the fingerprint representations were kept frozen. 
For cross-modal reconstruction tasks, only the corresponding decoder was trained. For instance, in the ECG-to-PPG reconstruction task, fingerprints were derived exclusively from ECG signals using the frozen ECG encoder from the foundation model, while the PPG decoder was trained to reconstruct the corresponding PPG signals.

For other downstream tasks, we employed e\textit{X}treme \textit{G}radient \textit{B}oosting (XGBoost). The model was trained with a learning rate of $0.1$, maximum tree depth of $6$, subsample and column-sampling ratios of $0.8$, and a minimum child weight of $1$. Up to $1{,}000$ boosting rounds were performed with early stopping (patience $= 50$ rounds) based on validation set performance. The objective function was selected according to task type: \texttt{binary:logistic} for binary classification tasks (mortality prediction, hypertension detection, and gender identification), \texttt{multi:softprob} for multi-class classification (CVD classification), and \texttt{reg:squarederror} for regression (age identification). GPU-accelerated training was enabled via the \texttt{tree\_method=gpu\_hist} parameter.

\subsection*{Data and pre-processing}

We employed large-scale waveform datasets for cross-modal foundation model training, as summarized in Table \ref{tab:data}. Cross-modal pre-training requires paired ECG and PPG recordings; therefore, we used MIMIC-III-WDB and VitalDB, from which we extracted 10-second paired segments. All signal segments were resampled to 2048 time points with minimal preprocessing. To broaden application scenarios and reduce the burden of data collection, we used only 10-second segments for both model training and fingerprint extraction. Among standard limb leads of ECG, lead II is widely recognized for providing the clearest and most diagnostically useful representation of P waves, QRS complexes, and T waves, essential for arrhythmia detection and heart rate variability analysis. Moreover, it is typically collected in various clinical and portable ECG devices \cite{perez2019large, goldberger2023goldberger, ramkumar2018atrial}. Therefore, to further emulate a wearable-device setting where typically only a single ECG lead is available, we utilized lead II ECG exclusively throughout this study, following practices in the existing literature \cite{hannun2019cardiologist}.

CODE-100\% and PTB-XL are ECG-only datasets and were not used for cross-modal pre-training; instead, they were used in downstream ECG-only evaluations.

The VitalDB and MIMIC data used in this study were obtained from the post-processed PulseDB. Raw ECG signals from PTB-XL and CODE datasets were processed using the \textit{ecg\_clean} function from the NeuroKit2 Python library \cite{makowski2021neurokit2} with default settings. All ECG and PPG segments were standardized to zero-mean and unit-variance.
To mitigate dataset-specific bias across heterogeneous cohorts, all samples from the merged datasets were jointly and randomly shuffled prior to training. This prevents the model from exploiting cohort-specific artifacts (\textit{e.g.}, acquisition protocols, device characteristics, or population prevalence rates) and encourages the learning of dataset-invariant, physiologically meaningful representations.

 Importantly, 20\% of each dataset listed in Table \ref{tab:data} was held out for testing and downstream tasks.

\subsection*{Evaluation methods}
We designed a range of downstream tasks to evaluate the effectiveness of the proposed biosignal fingerprints, including waveform reconstruction, binary classification, multi-class classification, and regression. For the reconstruction tasks, in addition to qualitative inspection, we employed the MAE Loss as the quantitative evaluation metric. For other types of tasks, the corresponding evaluation metrics are summarized in Table \ref{tab:metrics}.

\begin{table}[htbp]
\caption{\textbf{Supported evaluation metrics for different task types.}}
\label{tab:metrics}
\centering
\begin{tabular}{lcccccccccccc}
\toprule
\textbf{Task Type} & MAE & RMSE & $R^{2}$ & Pearson & AUROC & AUPRC & F1 & Accuracy & Precision & Recall \\
\midrule
Regression   & \checkmark & \checkmark & \checkmark & \checkmark &                         &             &          &           &        &                      \\
Binary       &            &            &            &                     & \checkmark  & \checkmark  & \checkmark  & \checkmark & \checkmark & \checkmark     \\
Multiclass   &            &            &            &                       & \checkmark  & \checkmark  & \checkmark  & \checkmark & \checkmark & \checkmark          \\
\bottomrule
\end{tabular}

\end{table}

We compared our fingerprints with six state-of-the-art foundation models, and SimCLR \cite{chen2020simclr}, NormWear \cite{luo2025foundationmodelmultivariatewearable}, PaPaGei \cite{pillai2025papageiopenfoundationmodels}, ECG-FM \cite{mckeen2024ecgfmopenelectrocardiogramfoundation}, MOMENT \cite{goswami2024moment}, and Chronos \cite{ansari2024chronos}. 

SimCLR is a landmark framework that established contrastive learning as a powerful paradigm for self-supervised representation learning by contrasting positive pairs (different augmentations of the same data) against negative pairs \cite{chen2020simclr}.
NormWear is a multi-modal foundation model designed to extract generalized representations from wearable sensor data including ECG and PPG, employing a channel-aware attention mechanism to capture both intra-signal and inter-signal relationships \cite{luo2025foundationmodelmultivariatewearable}. Pre-trained across various public wearable sensing datasets, it demonstrates strong performance and exceptional generalization capabilities across multiple downstream tasks.
PaPaGei is a foundation model for PPG signals that leverages domain knowledge of PPG signal morphology to achieve state-of-the-art results while being more data- and parameter-efficient \cite{pillai2025papageiopenfoundationmodels}. ECG-FM employs a Transformer-based architecture with pre-training using augmentation and contrastive learning tailored for ECG signals, demonstrating strong performance and reliable interpretability \cite{mckeen2024ecgfmopenelectrocardiogramfoundation}.
MOMENT is a general-purpose time series foundation model pre-trained on diverse time series data using masked autoencoding \cite{goswami2024moment}. Its unified Transformer architecture enables robust performance across forecasting, classification, anomaly detection, and imputation tasks with minimal fine-tuning.
Chronos treats time series forecasting as a language modeling task by tokenizing time series values and training autoregressive Transformers on this representation \cite{ansari2024chronos}. This approach achieves strong zero-shot forecasting capabilities and generalization to new domains without requiring task-specific fine-tuning.

Specifically, for the NormWear model, PPG and ECG signals are simultaneously input into the model, and the embeddings obtained through aggregation across patches are 768-dimensional, serving as the input for the classification model (XGBoost). On the other hand, PaPaGei and ECG-FM independently perform feature extraction on PPG and ECG signals, respectively, with embedding dimensions of 512 and 768, respectively. We used SimCLR-trained encoders to extract frozen embeddings as inputs to downstream models, matching the linear-probing evaluation protocol used for the other baselines. For downstream tasks requiring both PPG and ECG modalities, the concatenated embeddings from both PaPaGei and ECG-FM models (1280-dimensional) are used as the input for classification.

\section*{Data Availability}
The pre-training datasets PulseDB\footnote{https://github.com/pulselabteam/PulseDB}, which includes VitalDB and MIMIC-III WDB, and PTB-XL\footnote{https://physionet.org/content/ptb-xl/1.0.3/} are publicly available. For the access to CODE-100\%, please contact Antonio H. Ribeiro (antonio-ribeiro@ufmg.br) and Antonio Luiz P. Ribeiro (antonio.ribeiro@ebserh.gov.br) for more details.
The downstream datasets MIMIC-III\footnote{https://physionet.org/content/mimiciii/1.4/} and CODE15\footnote{https://zenodo.org/records/4916206} are publicly available online.

\section*{Acknowledgements}
We would like to express our gratitude to Antonio H. Ribeiro and Antonio Luiz P. Ribeiro for sharing the CODE-100\% dataset.
X.G. was supported by a Smart Data UK Research Fellowship (UKRI4004) and Medical and Life Science Translational Fund of University of Oxford (EP/X525777/1, MR/X50273X/1). D.A.C. was supported by the Pandemic Sciences Institute at the University of Oxford; the National Institute for Health Research (NIHR) Oxford Biomedical Research Centre (BRC); an NIHR Research Professorship; a Royal Academy of Engineering Research Chair; the Wellcome Trust funded VITAL project (grant 204904/Z/16/Z); the EPSRC (grant EP/W031744/1); and the InnoHK Hong Kong Centre for Cerebro-cardiovascular Engineering (COCHE). 

\section*{Author contributions statement}

Z.L. conceived the study, designed the methodology, performed the experiments, analyzed the results, prepared the figures, and wrote the manuscript. 
C.L. contributed to data preparation, performed the experiments, analyzed the results, and contributed to manuscript writing. 
F.L., Y.C., Y.Y., D.A.C., X.G. contributed to methodology design, result interpretation, manuscript revision, and finalization. 
Y.Y., D.A.C., X.G. supervised the project and revised the manuscript. 
All authors reviewed and approved the final manuscript.

\section*{Competing Interests}
The authors declare no competing interests.


\bibliography{sample}

\newpage
\appendix
\renewcommand{\thesection}{S\arabic{section}}
\setcounter{section}{0}

\appendix
\renewcommand{\thesection}{S\arabic{section}}
\setcounter{section}{0}

\renewcommand{\thefigure}{S\arabic{figure}}
\setcounter{figure}{0}

\renewcommand{\thetable}{S\arabic{table}}
\setcounter{table}{0}

\renewcommand{\theequation}{S\arabic{equation}}
\setcounter{equation}{0}

\section{Method-related Supplementary Material}

\subsection{Algorithm of signal augmentation strategy} \label{app:algo}

The signal augmentation strategy applied during cross-modal contrastive learning (Fig.~\ref{fig:method}B) is described in Algorithm~\ref{algo:1}.

\begin{algorithm}[H]
\caption{Signal Augmentation Strategy} \label{algo:1}
\begin{algorithmic}[1]
\Function{AugmentSignal}{$\text{signal}$}
    \State signal\_length $\gets$ length of signal
    \State time\_stamps $\gets \text{arange}(0, \text{signal\_length})$
    \State random\_curves $\gets \text{cumsum}(\mathcal{N}(0, 0.2, \text{signal\_length}))$
    \State random\_curves $\gets \frac{\text{random\_curves} - \min}{\max - \min}$
    \State warped\_time\_stamps $\gets$ time\_stamps + random\_curves

    \State \textit{Add white noise}: $\text{signal} \gets \text{signal} + \text{augment\_noise}$
    \State \textit{Time warping via interpolation:}
    \State \hskip1em $\text{warped\_signal} \gets \text{Interpolate}(\text{time\_stamps}, \text{warped\_time\_stamps}, \text{signal})$
    \State \textit{Normalize:}
    \State \hskip1em $\text{normalized\_signal} \gets (\text{warped\_signal} - \text{mean}) / \text{std\_dev}$
    \State \Return $\text{normalized\_signal}$ 
\EndFunction
\vskip 1em

\Function{GetAugmentIndices}{anchor\_signal}
    \State index\_list $\gets$  get all signal indices belonging to the same subject for a given $anchor\_signal$ 
    \If{length of index\_list > 3}
        \State augment\_indices =  Select two random indices;
    \ElsIf{length of index\_list == 2}
        \State augment\_indices $\gets$ index\_list        
    \Else \quad (a subject only has one segment of signal)
        \State  augment\_indices $\gets$ [index\_list[0], index\_list[0]]
    \EndIf
\State \Return  $\text{augment\_indices}$
\EndFunction

\vskip 1em
\Procedure{GenerateAugmentedPair}{$anchor\_index$, $ecg\_data$, $ppg\_data$}
    \State \textit{Get ECG anchor signal:} ecg\_anchor $\gets$  $ecg\_data$[$anchor\_index$]
    \State \textit{Get PPG anchor signal:} ppg\_anchor $\gets$  $ppg\_data$[$anchor\_index$]
    
    \State $i_{ecg} \gets$ \Call{GetAugmentIndices}{$ecg\_anchor$}
    \State $i_{ppg} \gets$ \Call{GetAugmentIndices}{$ppg\_anchor$}

    \If{$i_{ecg} = i_{ppg} = anchor\_index$}
        \State $ecg\_aug \gets$ \Call{AugmentSignal}{$ecg\_aug$}
        \State $ppg\_aug \gets$ \Call{AugmentSignal}{$ppg\_aug$}
    \ElsIf{$i_{ecg} = i_{ppg} \ne anchor\_index$}
        \If{Uniform$(0,1) < 0.5$}
            \State $ecg\_aug \gets$ \Call{AugmentSignal}{$ecg\_aug$}
        \Else
            \State $ppg\_aug \gets$ \Call{AugmentSignal}{$ppg\_aug$}
        \EndIf
    \Else
        \State $ecg\_aug \gets ecg\_data[i_{ecg}, :]$
        \State $ppg\_aug \gets ppg\_data[i_{ppg}, :]$
        \EndIf
    \State \Return ecg\_aug, ppg\_aug
\EndProcedure
\end{algorithmic}
\end{algorithm}

\subsection{Single-modal warm-start ablation}\label{app.warmstart}
\begin{figure}[h!]
\centering
\includegraphics[width=\linewidth]{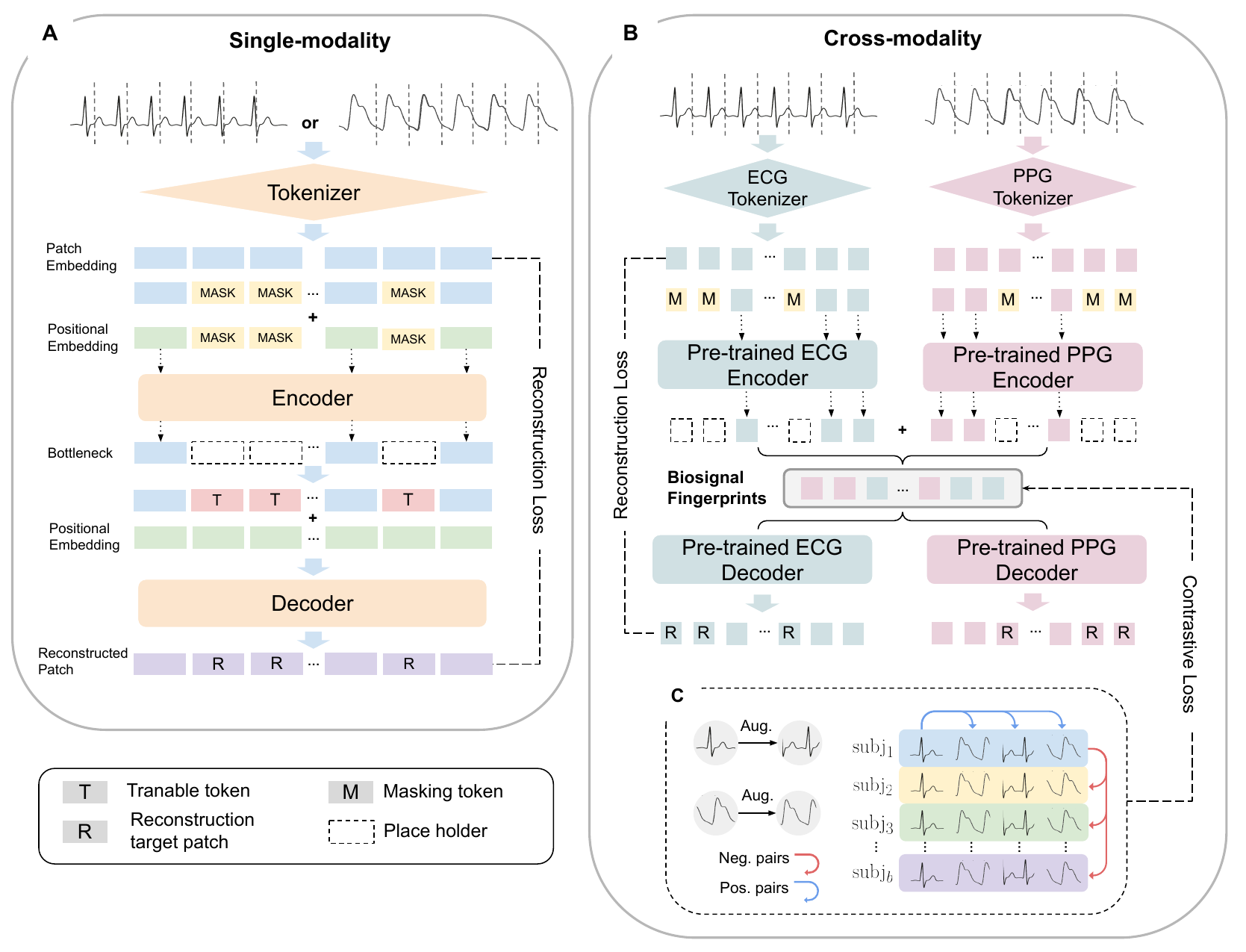}
\caption{\textbf{Two-stage Method Overview.} \textbf{A}: Illustration of the training framework for the single-modal models. The architecture follows a masked autoencoder framework built upon Vision Transformers (ViT). Two single-modal models were pre-trained on either ECG or PPG data to capture modality-specific representations. \textbf{B}: Illustration of the cross-modality training. This architecture is the same with the main pipeline as illustrated in Fig.~\ref{fig:method}. The only difference is that the encoders and decoders are initialized with the pre-trained components from Panel A. \textbf{C}: Overview of the data augmentation strategy for contrastive learning. This is also the same with Fig.~\ref{fig:method}.}
\label{appfig:method}
\end{figure}

We evaluated an optional warm-start strategy in which two single-modal masked autoencoders (ECG-only and PPG-only) were trained first, and their encoder/decoder weights were then used to initialize the cross-modal foundation model. 
We trained two single-modal models, one for ECG and one for PPG. These two single-modal pre-trained models were further served as initializations to the cross-modal training. The method illustration is shown in Fig.~\ref{appfig:method}. We experimented with different hyperparameter settings similar to the cross-modal training. The pre-training results for this ablation study are shown in Table \ref{apptab:ablation}. None of the settings outperform the single-stage cross-modal training. We therefore report the two-stage setup as an ablation and use the single-stage cross-modal pre-training protocol in the main paper.

\subsubsection{Single-modal training methods}
The input to the model encoder, $E_I$, was formed by the sum of the initial embedding matrix and the positional embeddings, $E_I=E + P_{\text{enc}}$. We then randomly masked a proportion $r_m$ of the patches. The resulting patch embeddings become:
\[
E_{\text{M}}[k_i] =
\begin{cases}
    E_{\text{I}}[k_i], & k_i \in U \quad (\text{unmasked row indices}), \\
    \text{[MASK]}, & k_i \in M \quad (\text{masked row indices}).
\end{cases}
\]
During the single-modal training stage, we followed the practice in the original work \cite{geng2022multimodal}, using 50\% as a masking ratio to better capture the global contextual information while facilitating the cross-model training next. The unmasked patch embeddings were passed to a Transformer encoder: $Z = f_{\text{enc}}(E_{\text{M}})$.

The final layers in the Transformer encoder are a dense output head, which generates the bottleneck embeddings of the M2AE, denoted as $Z$. 

The decoding phase first aligned the encoded unmasked patches with their original positions and filled the masked positions with trainable tokens,
\[
\widehat{Z}[k_i] =
\begin{cases}
    Z[k_i], & {k_i} \in U, \\
    T, & {k_i} \in M,
\end{cases}
\]
where $T\in \mathbb{R}^{d_{T}}$ are trainable position holding tokens. Mirroring the encoding phase, we added the positional embeddings $P_{\text{dec}}$ back to the encoded embeddings, $\widehat{Z} = \widehat{Z}+P_{\text{dec}}$. Finally, the recovered bottleneck $\widehat{Z}$ was fed into the decoder, another Transformer, for reconstruction:
$\widetilde{E} = f_{\text{dec}}(\widehat{Z})$.

The reconstruction loss was the mean absolute error (MAE Loss), computed only at the masked positions of the initial ``patchified'' embeddings $E$ of $x_i$.
\begin{equation}
 \mathcal{L}_{\text{recon}} = \frac{1}{|M|} \sum_{k_i \in M} \| E[k_i] - \widetilde{E}[k_i] \|_2^2.   
\end{equation}

\subsection{Hyperparameters tuning results for the single- and cross-modal foundation model training}\label{app.2}

\begin{table}[h]
\centering
\caption{\textbf{Final hyperparameters used in cross-modal training and the single-modal warm-start ablation study.}}\label{app.1}
\begin{tabular}{lrrr}
\toprule
Hyperparameters         & Single-modality ablation & Cross-modality training \\ \midrule
dropout                 & 0.1                      & 0.1                   \\
masking ts ratio      & 0.5                      & random {[}0.1-0.9{]}    \\
batch size             & 32                       & 256                     \\
learning rate          & $1e-3$                     & $1e-4$           \\
$\tau$             & -                        & 0.1                    \\
$\lambda$                  & -                        & 0.1                      \\ \bottomrule
\end{tabular}
\end{table}

\begin{table}[htbp]
\caption{\textbf{Hyperparameter tuning and corresponding loss metrics for the single-modal ablation study.} $\tau$ is the contrastive temperature; $\lambda$ is the reconstruction loss weight; $\mathcal{L}_{\text{contrast}}$ represents the conservative loss; $\mathcal{L}_{\text{recon}}^{\text{ECG}}$ and $\mathcal{L}_{\text{recon}}^{\text{PPG}}$ are reconstruction loss for ECG and PPG, respectively; \textit{random mask} indicates whether the model adopted random masking ratio and \textit{augmentation} indicates whether the model employed the positive pair augmentation.} \label{apptab:ablation}
\centering
\rowcolors{2}{white}{gray!10}
\begin{tabular}{cccccccccc}
\toprule
\textbf{batch size} & \textbf{$\tau$} & \textbf{$\lambda$} & \textbf{learning rate} & \textbf{total loss} & \textbf{$\mathcal{L}_{\text{contrast}}$} & \textbf{$\mathcal{L}_{\text{recon}}^{\text{ECG}}$}  & \textbf{$\mathcal{L}_{\text{recon}}^{\text{PPG}}$}  & \textbf{random mask} & \textbf{augmentation} \\
\midrule
256  & 0.1 & 0.1 & 1.00E-03 & 6.624 & 6.589 & 0.248  & 0.105 & TRUE & TRUE \\
256  & 0.1 & 0.1 & 1.00E-04 & 6.357  & 6.332 & 0.194 & 0.098 & TRUE  & TRUE  \\
256  & 0.1 & 0.01 & 1.00E-04 & 6.407  & 6.404 & 0.236 & 0.110 & TRUE  & TRUE  \\
256  & 0.1 & 0.1   & 1.00E-05 & 6.854  & 6.813 & 0.275 & 0.133 & TRUE  & TRUE  \\
1024 & 0.1 & 0.1 & 1.00E-04 & 6.372    & 6.343 & 0.203  & 0.093  & TRUE  & TRUE  \\
1024 & 0.05 & 0.1 & 1.00E-04 & 6.401  & 6.372 & 0.202 & 0.093 & TRUE  & TRUE  \\
\bottomrule
\end{tabular}

\end{table}

\begin{table}[h!]
\caption{\textbf{Hyperparameter settings and corresponding loss metrics for cross-modal training.} $\tau$ is the contrastive temperature; $\lambda$ is the reconstruction loss weight; $\mathcal{L}_{\text{contrast}}$ represents the conservative loss; $\mathcal{L}_{\text{recon}}^{\text{ECG}}$ and $\mathcal{L}_{\text{recon}}^{\text{PPG}}$ are reconstruction loss for ECG and PPG, respectively; \textit{random mask} indicates whether the model adopted random masking ratio and \textit{augmentation} indicates whether the model employed the positive pair augmentation. The highlighted row is the setting with the best performance and was adopted in the cross-modal foundation model.} \label{apptab:hypetune}
\centering
\rowcolors{2}{white}{gray!10}
\begin{tabular}{cccccccccc}
\toprule
\textbf{batch size} & \textbf{$\tau$} & \textbf{$\lambda$} & \textbf{learning rate} & \textbf{total loss} & \textbf{$\mathcal{L}_{\text{contrast}}$} & \textbf{$\mathcal{L}_{\text{recon}}^{\text{ECG}}$}  & \textbf{$\mathcal{L}_{\text{recon}}^{\text{PPG}}$}  & \textbf{random mask} & \textbf{augmentation} \\
\midrule
256  & 0.5 & 0.1 & 1.00E-03 & 7.4564 & 7.432 & 0.54  & 0.477 & FALSE & FALSE \\
\rowcolor{yellow}
256  & 0.1 & 0.1 & 1.00E-04 & 5.176  & 5.144 & 0.224 & 0.091 & TRUE  & TRUE  \\
256  & 0.5 & 0.1 & 1.00E-04 & 6.025  & 5.999 & 0.186 & 0.082 & TRUE  & TRUE  \\
256  & 0.5 & 1   & 1.00E-04 & 6.212  & 5.977 & 0.161 & 0.074 & TRUE  & TRUE  \\
512  & 0.1 & 0.1 & 1.00E-04 & 5.189  & 5.154 & 0.24  & 0.108 & TRUE  & TRUE  \\
1024 & 0.1 & 0.1 & 1.00E-04 & 5.200    & 5.164 & 0.25  & 0.110  & TRUE  & TRUE  \\
1024 & 0.5 & 0.1 & 1.00E-04 & 6.024  & 5.989 & 0.227 & 0.102 & TRUE  & TRUE  \\
1024 & 0.5 & 1   & 1.00E-04 & 6.238  & 5.981 & 0.177 & 0.080 & TRUE  & TRUE  \\
1024 & 0.1 & 0.1 & 1.00E-03 & 6.356  & 6.314 & 0.312 & 0.145 & TRUE  & TRUE  \\
\bottomrule
\end{tabular}

\end{table}

\newpage

\section{Result-related Supplementary Material}
\subsection{Qualitative results for hypertension classification}

\begin{figure}[h!]
\centering
    \centering
    \includegraphics[width=0.7\linewidth]{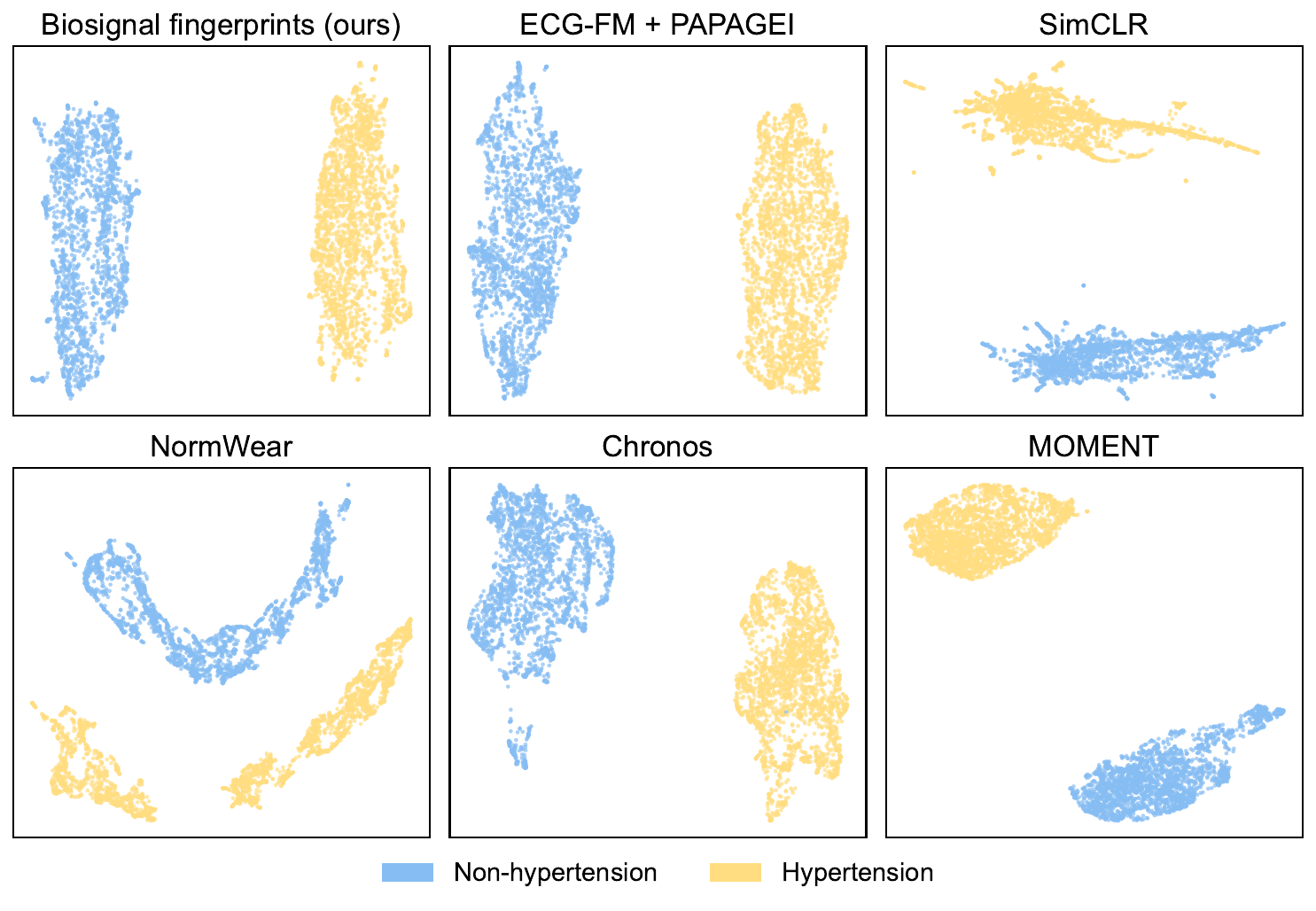}
 \caption{\textbf{UMAP projections of embeddings generated by the six foundation models and colored by class labels for the hypertension task.} }\label{appfig:umap_hyp}
\end{figure}

To complement the quantitative results of the hypertension classification task, we performed the UMAP visualization - projecting the latent representations from all methods to a 2-dimensional space as shown in Fig.~\ref{appfig:umap_hyp}. UMAP projections reveal that while all six models achieve visible separation between hypertensive and non-hypertensive individuals, the quality of that separation differs markedly. Biosignal fingerprints produce two compact, well-bounded clusters with clear inter-class margins. ECG-FM + PaPaGei and SimCLR yield broader, more diffuse distributions with greater intra-class scatter, while NormWear generates fragmented sub-clusters suggesting its representations encode structure beyond hypertension status. Chronos achieves reasonable inter-class separation but with less compact cluster geometry than biosignal fingerprints. MOMENT produces visually distinct blobs that nonetheless fail to support competitive linear classification, likely reflecting a less amenable embedding geometry in the full high-dimensional space. 

\subsection{Quantitative results for all classification tasks}\label{app.3}
\begin{table}[h!]
\caption{\textbf{Results of downstream classification tasks.} AUPRC represents Area Under the Precision-Recall Curve, and AUROC represents Area Under the Receiver Operating Characteristic Curve. Our proposed method is highlighted in bold font. }
\begin{tabular}{llcccccc}
\toprule
\textbf{Task}                 & \textbf{Method}               & \textbf{AUPRC} & \textbf{AUROC} & \textbf{Accuracy} & \textbf{F1-score} & \textbf{Precision} & \textbf{Recall} \\ \midrule

\multirow{6}{*}{CVD 5-class}          & \textbf{Biosignal fingerprints (ours)} & \textbf{0.920}  & \textbf{0.974} & \textbf{0.856}    & \textbf{0.855}    & \textbf{0.856}     & \textbf{0.856}  \\
                              & SimCLR            & 0.733             & 0.905            & 0.684  & 0.673       & 0.677       & 0.674       \\
                              & ECG-FM             &     0.931           &    0.978            &  0.869 &       0.861      &       0.864       &         0.861       \\
                              & NormWear        &      0.758      &      0.913      &      0.696     &     0.681         &     0.684       &     0.682       \\
                              &Chronos & 0.865 &0.955 &0.807 &0.793 &0.798 &0.794\\ 
                              & MOMENT &0.586 &0.834 &0.555 &0.524 &0.527 &0.531\\\midrule
\multirow{6}{*}{CVD 3-class}          & \textbf{Biosignal fingerprints (ours)} & \textbf{0.901}  & \textbf{0.956} & \textbf{0.847}    & \textbf{0.760}    & \textbf{0.910}     & \textbf{0.705}  \\
                              &SimCLR            &  0.869                 &        0.930          &  0.835      &     0.756        &      0.918       &    0.690          \\
                              & PaPaGei & 0.623 & 0.759 & 0.688 & 0.438 & 0.792 & 0.436 \\
                              & NormWear & 0.758 & 0.871 & 0.767 & 0.620 & 0.768 & 0.580 \\
                              & Chronos & 0.793 & 0.880 & 0.773 & 0.592 & 0.887 & 0.571 \\
                              & MOMENT & 0.487 & 0.679 & 0.653 & 0.369 & 0.399 & 0.394 \\\midrule

\multirow{6}{*}{Mortality}    & \textbf{Biosignal fingerprints (ours)} & \textbf{0.693} & \textbf{0.729} & \textbf{0.676}     & \textbf{0.676}    & \textbf{0.676}      & \textbf{0.676}   \\
                              & SimCLR            & 0.663             & 0.678            & 0.630  & 0.629       & 0.629       & 0.629       \\
                              & ECG-FM     &  0.682    &     0.713   &  0.657     &    0.657   &    0.657    &    0.657   \\
                              & NormWear    &   0.651    &   0.674 &    0.632        &    0.632      &      0.632    &   0.632              \\ 
                              & Chronos & 0.667 &0.699 &0.651 &0.650 &0.651 &0.650\\
                              & MOMENT & 0.636 & 0.657 &0.604 &0.604 & 0.604 &0.604\\\midrule
\multirow{6}{*}{Hypertension} & \textbf{Biosignal fingerprints (ours)} & \textbf{0.882} & \textbf{0.877} & \textbf{0.787}    & \textbf{0.786}    & \textbf{0.788}     & \textbf{0.786}  \\
                              & SimCLR            & 0.865             & 0.857            & 0.765  & 0.764       & 0.765       & 0.764       \\
                              & ECG-FM \& PaPaGei             & 0.803          & 0.793          & 0.707             & 0.706             & 0.707              & 0.706           \\
                              & NormWear                      & 0.755          & 0.746          & 0.670              & 0.668             & 0.670               & 0.668           \\
                              & Chronos &0.704 &0.704 &0.647 &0.645 &0.647 &0.646\\
                              & MOMENT &0.637 &0.636 &0.591 &0.596 &0.596 &0.593\\\midrule
\multirow{6}{*}{Gender} & \textbf{Biosignal fingerprints (ours)} & \textbf{0.889} & \textbf{0.900} & \textbf{0.813}    & \textbf{0.812}    & \textbf{0.812}     & \textbf{0.812}  \\
                              & SimCLR            & 0.869             & 0.880            & 0.794  & 0.793       & 0.793       & 0.792       \\
                              & ECG-FM \& PaPaGei         & 0.804	& 0.823	& 	0.737	& 0.735 & 	0.736 & 	0.734         \\
                              & NormWear      & 0.768 &	0.790	&	0.708	&  0.706 & 	0.707 & 0.706 \\ 
                              & Chronos &0.739 &0.770 &0.699 &0.696 &0.697 &0.695 \\
                              & MOMENT &0.612 &0.656 &0.610 &0.605 &0.608 &0.605\\
          \bottomrule

\end{tabular}
\end{table}
\newpage

\end{document}